\documentclass{article} 
\usepackage{iclr2027_conference,times}

\usepackage{amsmath,amsfonts,bm}

\def\eqref#1{equation~\ref{#1}}

\def\1{\bm{1}}

\DeclareMathAlphabet{\mathsfit}{\encodingdefault}{\sfdefault}{m}{sl}
\SetMathAlphabet{\mathsfit}{bold}{\encodingdefault}{\sfdefault}{bx}{n}

\usepackage{wrapfig}
\usepackage{amsmath,amssymb}
\usepackage{graphicx}
\usepackage{booktabs}
\usepackage{caption}
\usepackage{multirow}
\usepackage{array}
\usepackage{tabularx}
\usepackage{makecell}
\usepackage{siunitx}
\usepackage{algorithm}
\usepackage{algorithmic}
\usepackage{xcolor}
\usepackage{wrapfig}
\usepackage[section]{placeins}
\usepackage{hyperref}
\usepackage{url}
\usepackage{xcolor}
\usepackage{booktabs}
\usepackage{colortbl}
\definecolor{bestgreen}{RGB}{225,242,225}
\definecolor{nagray}{RGB}{246,246,246}
\definecolor{bestgreen}{RGB}{236,247,236}

\newcommand{\profilecell}[1]{\cellcolor{bestgreen}\textbf{#1}}
\newcommand{\bestcell}[1]{\cellcolor{bestgreen}\textbf{#1}}
\newcommand{\nacell}{\cellcolor{nagray}\textcolor{gray}{N/A}}
\newcommand{\cival}[2]{#1\,{\scriptsize$\pm$\,#2}}
\definecolor{temporalblue}{RGB}{223,233,246}
\definecolor{spatialorange}{RGB}{247,228,216}
\definecolor{spatiotemporalpink}{RGB}{236,208,236}

\newcommand{\memmark}[1]{%
  \raisebox{0.05ex}{%
    \colorbox{#1!85!white}{%
      \rule{0pt}{0.72em}\hspace{0.72em}%
    }%
  }\hspace{0.35em}%
}

\newcommand{\temporal}{%
  \memmark{temporalblue}Temporal}

\newcommand{\spatial}{%
  \memmark{spatialorange}Spatial}

\definecolor{LMActive}{HTML}{5F7FA3}
\definecolor{LMScene}{HTML}{C28A5B}
\definecolor{LMEntity}{HTML}{B77B8D}

\graphicspath{{figures/}}

\definecolor{SpatialColor}{HTML}{00796B}

\newcommand{\method}{GROME}

\title{\method: GROunding Factorized MEmory into Video World Models}

\author{%
\makebox[\textwidth][l]{%
\begin{tabular}{l}
{\normalfont\normalsize
Fang Yu$^{1,2,*}$,
Xiaoxue Wu$^{1,3}$,
Haiyu Zhang$^{4}$,
Yangfu Li$^{2}$,
Botong Zhao$^{2}$,
}\\[0.5ex]
{\normalfont\normalsize
Yu Qiao$^{1}$,
Tianfan Xue$^{1,5}$,
Yue Lu$^{2}$,
Xinyuan Chen$^{1,\dagger}$
}\\[1.5ex]
{\normalfont\footnotesize
$^{1}$Shanghai Artificial Intelligence Laboratory
}\\[0.4ex]
{\normalfont\footnotesize
$^{2}$East China Normal University
\qquad
$^{3}$Fudan University
}\\[0.4ex]
{\normalfont\footnotesize
$^{4}$Beihang University
\qquad
$^{5}$The Chinese University of Hong Kong
}
\end{tabular}%
}%
}

\iclrfinalcopy
\makeatletter
\def\lhead#1{}
\makeatother
\begin{document}

\maketitle
\lhead{}

\begingroup
\renewcommand{\thefootnote}{\fnsymbol{footnote}}
\footnotetext[1]{Work done during internship at Shanghai AI Laboratory.}
\footnotetext[2]{Corresponding author.}
\endgroup
\vspace{-3em}
\begin{center}
  \small
  \textbf{Project Page:}
  \url{https://lyeean.github.io/GROME_/}
\end{center}
\vspace{-1em}
\begin{figure}[htbp]
  \centering
  \includegraphics[width=\linewidth]{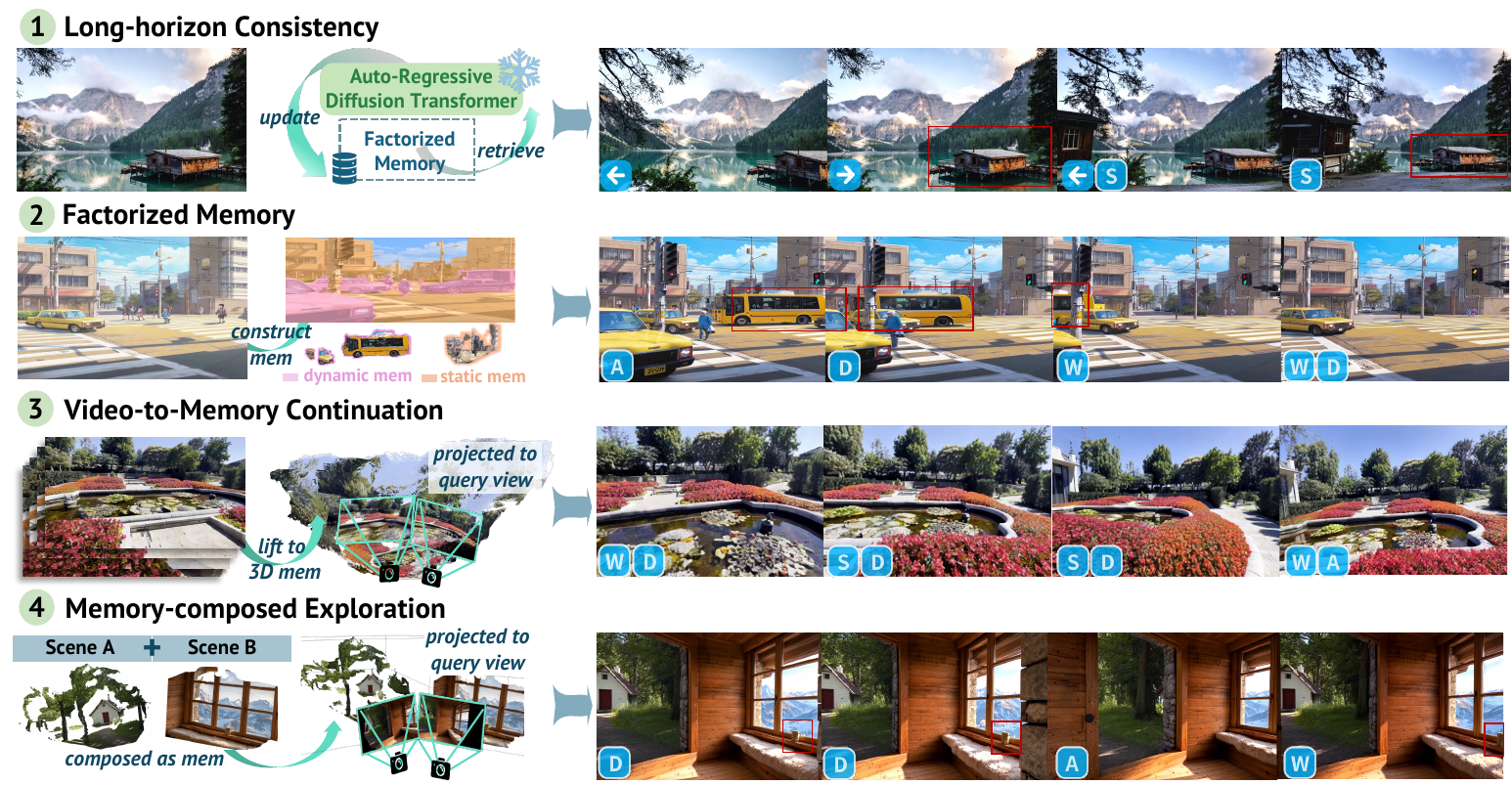}
  \caption{
    \textbf{(1)} 
    During autoregressive rollout, GROME progressively updates historical observations into a factorized memory and injects retrieved local evidence into a frozen video world model.
    \textbf{(2)} 
    By decoupling static scene structures from dynamic entities, it ensures scene persistence while accommodating entity dynamics.
    \textbf{(3)} 
    GROME also supports lifting real-world videos into factorized memory as plug-and-play references for controllable video continuation.
    \textbf{(4)} Multiple point clouds can be seamlessly composed as unified external memory for open-ended spatial exploration.}
  \label{fig:intro1}
\end{figure}
\begin{abstract}
  Long-horizon consistency is essential for video world models, yet it becomes difficult to maintain once observed content falls outside the context window.
  Existing memory mechanisms either retrieve history as implicit context, which remains constrained by context capacity and entangles static and dynamic information or consolidate history into explicit global representations, which struggle to model evolving dynamics and accumulate geometric errors. 
  To overcome these limitations, we introduce \textbf{GROME}, a novel factorized memory framework that augments video world models for long-horizon coherence across both static scenes and dynamic entities. 
  GROME operates via three synergistic components: 
  1)~\emph{Static-Dynamic Factorization} decouples historical evidence into static and dynamic components, preserving static scene geometry and dynamic entity states separately; 
  2)~\emph{Local Memory Retrieval} selects spatially aligned static history to eliminate geometric drift alongside the latest dynamic entity states to ensure temporal continuity;
  3)~\emph{Training-Free Memory Injection} seamlessly grounds the factorized memory into video world models via their native conditioning pathways.
  Experiments on long-horizon revisit scenarios demonstrate that GROME substantially enhances scene persistence and dynamic entity fidelity over existing methods, all while preserving the visual quality and controllability of the frozen backbone without additional training.
  Beyond self-rollouts, GROME readily accommodates external observations as plug-and-play memory, unlocking flexible and controllable video continuation and scene exploration.\end{abstract}

\begin{figure*}[h]
  \centering
  \includegraphics[width=\textwidth]{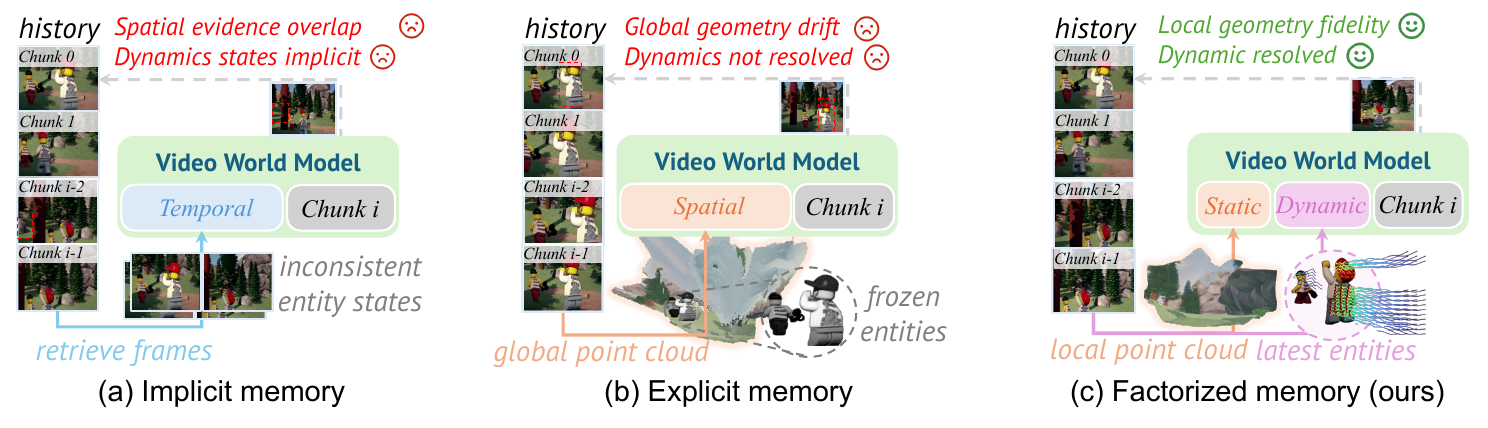}
  \caption{
    \textbf{Comparison of memory paradigms.}
    Implicit memory suffers from substantial scene-content overlap and ambiguous in dynamic states, while explicit memory consolidates historical frames into a global 3D scene but struggles with evolving dynamic entities and accumulated geometric drift.
    GROME instead factorizes historical frames into static-dynamic memory, preserving both persistent static scene structure and evolving dynamic entities over long-horizon generation.
    }
  \label{fig:intro}
\end{figure*}

\section{Introduction}

Long-horizon consistency is essential for video world models. It ensures that the model preserves coherent internal world states across control inputs, thereby synthesizing geometrically and temporally consistent visual frames. This capability is pivotal for downstream applications spanning interactive entertainment and embodied AI~\citep{zhu2024irasim,yang2023learning,hu2023gaia,zhou2024robodreamer,kanervisto2025world,he2026pre,tang2025hunyuan}.

Existing studies on modeling long-horizon consistency broadly fall into two paradigms as shown in Fig.~\ref{fig:intro}. Implicit memory methods~\citep{yu2025context, sun2025worldplay, li2025vmem} maintain historical observations and retrieve relevant context, encouraging the model to implicitly learn geometric consistency. However, these methods are constrained by finite context windows, hindering their ability to retain all historical information necessary for continuous world evolution. Moreover, disentangling static  content from dynamic variations solely from visual observations is inherently ill-posed and challenging, potentially inducing temporal inconsistencies over extended horizons. 

Alternatively, explicit memory methods~\citep{ren2025gen3c, wu2025spatialmemory, chen2025teleworld} reconstruct historical information into compact 3D representations, \textit{e.g.}, point clouds or 3D Gaussians, to sustain persistent memory. While these approaches model long-horizon geometric consistency more efficiently, they focus on reconstructing static scene content while neglecting dynamic world attributes. Consequently, they struggle to faithfully simulate time-varying environmental dynamics and entity motions. Moreover, due to compounding reconstruction errors and surface stratification artifacts over long horizons, maintaining a globally consistent explicit 3D representation becomes prohibitively difficult, ultimately failing to provide reliable guidance for subsequent world evolution. 
Therefore, video world models fundamentally require a coherent memory representation capable of preserving static scene content while adaptively retaining the evolving states of dynamic entities.

To bridge this gap, we introduce GROME, a novel factorized memory that explicitly disentangles static scene content from evolving dynamics, which can be injected into video world models in a training-free manner. Specifically, we propose \textbf{Static-Dynamic Factorization}, which performs per-chunk 3D reconstruction and decomposes the resulting pointmaps into static and dynamic components. To avoid the geometric drift caused by global consolidation, both types of components are stored as local pointmaps, preserving the geometric fidelity of observations over extended horizons. 
Under a bounded context budget, we present \textbf{Local Memory Retrieval}, which executes distinct retrieval strategies tailored to the characteristics of the two components. Static pointmaps are retrieved to maximize spatial coverage relative to target views and consolidated into reference views, thereby offering rich and consistent geometric context. 
To encourage the model to treat these references as invariant spatial anchors, we assign them the same temporal RoPE indices as the target views, effectively enforcing spatial consistency. In contrast, dynamic pointmaps are ranked by temporal recency, with only the latest dynamic evidence selected as the source for rendering target-view references. Here, we preserve their original temporal RoPE indices to maintain the causal temporal dynamics. 

Furthermore, we introduce \textbf{Training-Free Memory Injection} to seamlessly integrate our factorized memory into video world models via their native conditioning pathways. Since our rendered reference inevitably contain unobserved and occluded regions, these contexts induces a noticeable distributional mismatch with the backbone’s native feature space. Therefore, we employ \textit{Reference Value Adaptation}, which dynamically modulates the rendered contexts within attention blocks, aligning their numerical statistics with the pretrained feature manifold. To prevent mutual interference between the static and dynamic streams, we design \textit{Static-Dynamic Disentangled Attention} to register both memory contexts through separate attention pathways. Finally, we incorporate \textit{Timestep-Aware Memory Routing} to dynamically schedule the relative influence of both memory components across diffusion denoising timesteps, enabling effective coarse-to-fine memory conditioning.

The resulting system equips video world models with a factorized memory mechanism for long-horizon autoregressive generation, enabling them to preserve static scene content and evolving dynamic entity states beyond the model's finite context window without fine-tuning.
Experiments on long-horizon revisit scenarios involving both static and dynamic scenes demonstrate the effectiveness of GROME. 
Across multiple LingBotV2 variants and representative memory-based baselines, GROME consistently improves revisit consistency and dynamic entity preservation while retaining the visual quality and controllability of the frozen backbone. 
Ablation studies further confirm that static-dynamic factorization, local memory retrieval, and training-free memory injection provide complementary benefits, together accounting for the improvements in long-horizon consistency.

\section{Related Work}
\label{sec:related_work}
\textbf{Interactive Video World Models}.
Interactive video world models predict future visual observations conditioned on user actions~\citep{wu2024ivideogpt,wang2023drivedreamer,wang2023drivewm}. Recent generative video world models have substantially improved visual fidelity and visual environment diversity~\citep{agarwal2025cosmos,ni2025maskgwm,yu2025gamefactory,guo2025mineworld}. Meanwhile, contemporary systems increasingly support real-time user exploration and richer multi-modal control~\citep{he2025matrixgame2,mao2025yume,yan2025foundational,gao2026infinite,pan2025worldmodel,liao2025genieenvisioner}. Despite these remarkable advances, maintaining long-horizon consistency remains a critical challenge: under continuous action sequences, previously observed scene structure and evolving dynamic states inevitably degrade due to catastrophic forgetting once they exit the finite context window. To bridge this gap, GROME directly overcomes this limitation by augmenting frozen video world models with factorized static–dynamic memory in a training-free manner, faithfully preserving both persistent 3D structure and dynamic entities simultaneously over extended rollouts.

\textbf{Memory for Long-Horizon World Modeling}.
To maintain long-horizon consistency beyond the finite context window, recent video world models increasingly introduce different memory mechanisms to preserve historical evidence. Existing approaches broadly follow two paradigms: implicit memory, which retrieves historical observations as model context, and explicit memory, which reconstructs history into persistent structured representations.
Implicit memory methods directly reuse historical observations~\citep{yu2025context,li2025vmem}; related methods further retrieve past observations or learned memory to extend effective context~\citep{xiao2025worldmem,yu2026memlearner,sun2025worldplay,li2025gamecraft}.
These approaches retain direct visual evidence and improve access to history, but remain fundamentally bounded by the model's finite context capacity, making it difficult to preserve all historical information required for continuous world evolution.
Moreover, static scene structure and dynamic variations remain entangled in the retrieved visual context, leaving the model to implicitly infer which content should remain persistent and which should evolve over time. This ambiguity becomes increasingly difficult to resolve over long rollouts and can lead to temporal inconsistencies.
Explicit memory methods~\citep{team2026alayaworld,Wang2026WarpasHistoryGC,yu2026mosaicmem} instead reconstruct history into persistent geometric representations.
Gen3C~\citep{ren2025gen3c} maintains a global 3D cache, while broader variants maintain geometric memory in 3D or latent spaces
~\citep{wu2025spatialmemory,wang2026latent}.
These representations provide more persistent geometric grounding, but largely emphasize stable scene content and remain challenged by evolving dynamic states~\citep{wang2026anchorweave,zhao2026spatia}.
To address evolving dynamic states and cumulative geometric drift, GROME factorizes historical evidence into static and dynamic local memories, preserving persistent scene content and time-varying entity states without global consolidation.
\section{Method}
\label{sec:method}
\begin{figure*}[t]
  \centering
  \includegraphics[width=\textwidth]{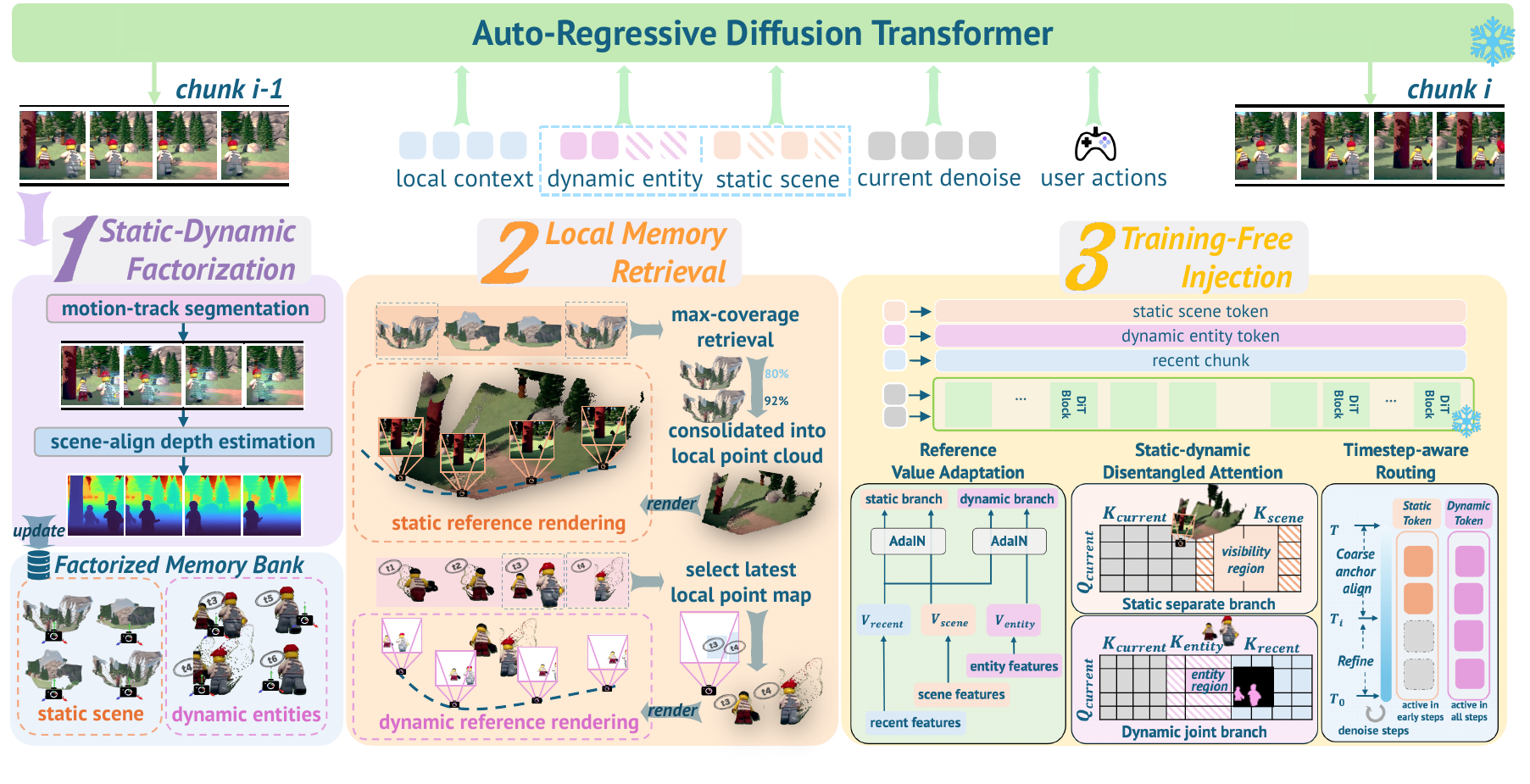}
  \caption{
    \textbf{Overview of GROME}. GROME factorizes historical evidence into static scene and dynamic entity components. Both are retained over extended rollouts as local spatial memories, with the dynamic component further organized by explicit temporal indexing. Given a target camera trajectory, relevant historical evidence is separately retrieved from the static and dynamic components to render references. These static-dynamic references are then injected into the model's native history-conditioning pathway through a set of training-free injection strategies, conditioning generation together with the local context and user actions. This allows GROME to better preserve consistency over long rollouts while maintaining generation quality and controllability.
      }
  \label{fig:method_overview}
\end{figure*}

This section presents the core design of \textbf{GROME} (Fig.~\ref{fig:method_overview}). We first formulate the problem by highlighting the bounded-context bottleneck in world models and our memory augmentation objective (Sec.~\ref{sec:method-overview}). We then introduce \textbf{Static-Dynamic Factorization} to decouple history (Sec.~\ref{sec:method-construction}), and execute \textbf{Local Memory Retrieval} via asymmetric policies (Sec.~\ref{sec:method-projection}). Finally, to condition the frozen backbone without retraining, we apply \textbf{Training-Free Memory Injection} (Sec.~\ref{sec:method-registration}). Together, these components ensure robust long-horizon consistency across rollouts.
\subsection{Problem Formulation}
\label{sec:method-overview}

We consider a pretrained, action-conditioned video world model $\mathcal F_\theta$ that autoregressively synthesizes future video rollouts. At step $t$, let $X_t$ denote the generated video chunk and $a_t$ the corresponding action conditioning (e.g., camera trajectory). Constrained by memory and computation, the frozen backbone retains only a bounded window of recent history:
\begin{equation}
\mathcal C_{t-1}
=
\left\{
X_i
\right\}_{i=\max(1,t-L_{\mathrm{ctx}})}^{t-1},
\qquad
|\mathcal C_{t-1}| \leq L_{\mathrm{ctx}},
\label{eq:active_context}
\end{equation}
where $L_{\mathrm{ctx}}$ specifies the context capacity of the pretrained backbone.

The world model maps this active context into its native historical condition $\mathcal R_t^{\mathrm{RC}}$ via an internal conditioning operator $\mathcal H$, generating the next chunk as:
\begin{equation}
\mathcal R_t^{\mathrm{RC}}
=
\mathcal H(\mathcal C_{t-1}),
\qquad
X_t
=
\mathcal F_\theta
\left(
a_t,
\mathcal R_t^{\mathrm{RC}}
\right),
\qquad
\theta\ \text{frozen}.
\label{eq:vanilla_rollout}
\end{equation}

During autoregressive rollout, observations older than $L_{\mathrm{ctx}}$ steps are inevitably evicted from $\mathcal C_{t-1}$. Consequently, $\mathcal R_t^{\mathrm{RC}}$ loses historical grounding over extended horizons, leading to catastrophic forgetting of previously visited environments and inconsistent dynamic entity evolution.
To overcome this bounded-context constraint without retraining the backbone, we formulate the objective as maintaining an external, persistent world memory $\mathcal M_{t-1}$ across rollouts. At each step $t$, we reallocate a portion of the bounded context budget in $\mathcal R_t^{\mathrm{RC}}$ to host relevant evidence retrieved from $\mathcal M_{t-1}$, thereby enabling long-horizon consistency within the native capacity of the frozen backbone $\mathcal F_\theta$.

\subsection{Static-Dynamic Factorization}
\label{sec:method-construction}

Static scene content remains largely persistent across observations, whereas
dynamic entities may occupy different positions and states over time.
Consequently, retaining only static content discards historical evidence of
dynamic entities, while directly consolidating dynamic observations into a
shared spatial representation can mix incompatible states from different
times. To preserve both types of historical evidence, GROME factorizes historical evidence into two components,
\begin{equation}
\mathcal M_{t-1}
=
\left(
\mathcal M_{t-1}^{S},
\mathcal M_{t-1}^{D}
\right),
\label{eq:static_dynamic_memory}
\end{equation}
where $\mathcal M_{t-1}^{S}$ denotes static memory and
$\mathcal M_{t-1}^{D}$ denotes dynamic memory.
Static scene content is retained as spatially addressable evidence, while
dynamic entities are stored separately with explicit temporal indexing.
In this way, GROME preserves both persistent scene content and the time-varying states of dynamic entities, preventing mutual interference throughout long-horizon rollouts.

A separate challenge arises from long-term geometric accumulation.
Depth estimated from generated observations inevitably contains
errors, and repeatedly consolidating these observations into a persistent
global point cloud can propagate such errors and progressively degrade the
geometry of rendered references.
Therefore, GROME retains both static and dynamic evidence as local
pointmaps. This preserves the local geometric fidelity of individual
observations while allowing them to be spatially composed only when required
by future generation.

\textbf{Local 3D Observation Lifting.}
For each historical chunk $X_i$, we first apply motion-aware
segmentation to separate static scene regions from dynamic content.
Let $X_{i,f}$ denote the $f$-th frame of chunk $X_i$.
For each frame, we estimate its depth map and align it to the existing
static memory using reliable static overlap.
The aligned depth $D_{i,f}$ is then back-projected into a local pointmap,
\begin{equation}
P_{i,f}
=
\Pi_D^{-1}
\left(
X_{i,f},
D_{i,f},
K_{i,f}
\right),
\label{eq:observation_lifting}
\end{equation}
where $K_{i,f}$ denotes the camera intrinsics.
The corresponding motion-aware segmentation further separates $P_{i,f}$ into
a static pointmap $P_{i,f}^{S}$ and a dynamic pointmap $P_{i,f}^{D}$.
Each pointmap remains local to its source observation rather than being permanently fused with previous observations.
Details of depth alignment and segmentation are provided in Appendix~\ref{app:depth_alignment} and Appendix~\ref{app:factorized_memory}.
Let $F_i$ denote the frame-index set of chunk $X_i$,
the two memory components retain local pointmaps with different indexing information:
\begin{equation}
\begin{aligned}
\mathcal M_{t-1}^{S}
&=
\left\{
(P_{i,f}^{S},T_{i,f})
\;\middle|\;
i<t,\;
f\in\mathcal F_i,\;
P_{i,f}^{S}\ \text{valid}
\right\},
\\
\mathcal M_{t-1}^{D}
&=
\left\{
(P_{i,f}^{D},T_{i,f},\tau_{i,f})
\;\middle|\;
i<t,\;
f\in\mathcal F_i,\;
P_{i,f}^{D}\ \text{valid}
\right\}.
\end{aligned}
\label{eq:factorized_memory}
\end{equation}

where \(T_{i,f}\) provides spatial addressability for both memory components, while \(\tau_{i,f}\) additionally records the source temporal position of dynamic evidence.

\subsection{Local Memory Retrieval}
\label{sec:method-projection}

Retaining historical observations as local pointmaps avoids cumulative
geometric drift, but causes the external memory to grow continuously over
long rollouts, while the conditioning budget of the frozen backbone remains
bounded. GROME therefore retrieves query-relevant evidence from the growing memory and renders references within a bounded context budget. Importantly, static and dynamic memories require different retrieval
criteria: static observations provide complementary spatial coverage, whereas
dynamic observations may represent competing states at different times.

\textbf{Distinct Memory Retrieval.}
Given the target camera trajectory $q_t$, GROME retrieves the two memory
components according to their distinct characteristics.
For static scene memory, we select complementary local pointmaps using a max-coverage criterion~\citep{wang2026anchorweave}, favoring observations that provide additional coverage of the target views. The selected pointmaps are temporarily
composed and projected to $q_t$, yielding a static reference with rich spatial coverage while retaining the geometric fidelity of the original local
observations.

For dynamic entity memory, only the latest dynamic evidence is selected as the source for target-view reference rendering.
We denote the resulting references as
\begin{equation}
\widetilde{\mathcal M}_{t}
=
\left(
\widetilde{\mathcal M}_{t}^{S},
\widetilde{\mathcal M}_{t}^{D}
\right),
\label{eq:memory_retrieval}
\end{equation}
where $\widetilde{\mathcal M}_{t}^{S}$ and
$\widetilde{\mathcal M}_{t}^{D}$ denote the retrieved static and dynamic references, respectively. Details of retrieval and geometric projection are
provided in Appendix~\ref{app:memory_projection}.

\textbf{Spatiotemporal RoPE.}
Both static and dynamic references are rendered into the target views and use query-aligned spatial RoPE indices, while their temporal indices
are assigned differently.
To encourage the model to treat static references as invariant spatial anchors,
we assign them the same temporal RoPE indices as the target views, following
\citet{Wang2026WarpasHistoryGC}.
Dynamic references, in contrast, are spatially aligned with the query views
but still represent entity states observed at their source times.
We therefore preserve their original temporal RoPE indices, preserving its causal temporal dynamics and preventing
historical entity states from being interpreted as current observations.

\subsection{Training-Free Memory Injection}preserving its causal temporal dynamics and preventing
\label{sec:method-registration}

To incorporate the retrieved static-dynamic references into a frozen world
model, GROME addresses three mismatches with the backbone's native
history-conditioning pathway: reference statistics, heterogeneous memory
semantics, and timestep-dependent conditioning requirements.

\textbf{Reference Value Adaptation.}
Although projection aligns historical evidence geometrically with the target
views, references retrieved from distant observations may contain partial
support and exhibit appearance statistics that differ from those of the recent rollout.
Directly injecting such references can therefore introduce a distributional
mismatch with the backbone's native context features.
GROME uses the retained recent context as an appearance anchor and aligns the
value statistics of each memory stream before attention. For
$m\in\{S,D\}$,
\begin{equation}
\widehat V^{m}
=
\frac{
V^{m}-\mu(V^{m})
}{
\sigma(V^{m})+\epsilon
}
\sigma(V^{\mathrm{ctx}})
+
\mu(V^{\mathrm{ctx}}),
\qquad
\bar V^{m}
=
(1-\lambda_m)V^{m}
+
\lambda_m\widehat V^{m},
\label{eq:value_alignment}
\end{equation}
where $\mu(\cdot)$ and $\sigma(\cdot)$ are computed over valid memory tokens,
and $\lambda_m$ controls the alignment strength of each stream.
Only the values are adapted, while the corresponding keys $K^m$ remain
unchanged, preserving the original query--key correspondence.

\textbf{Static-Dynamic Disentangled Attention.}
Static and dynamic references carry different spatial and temporal semantics:
static memory provides dense scene-level support, whereas dynamic memory
contains temporally indexed entity evidence.
Normalizing both streams within a single attention distribution would force
these heterogeneous references to compete for attention.
We therefore inject them through two independently normalized attention
branches. The dynamic branch jointly attends to current, recent-context, and
dynamic-memory tokens, while the static branch attends to current and
static-memory tokens, avoiding mutual interference between distinct memories.

Let $(K^m,V^m)$, $m\in\{\mathrm{cur},\mathrm{ctx},D,S\}$, denote the KV
states of current, recent-context, dynamic, and static tokens, respectively,
and let $Q^{\mathrm{cur}}$ denote the current-token queries.
Using the aligned memory values from Eq.(~\ref{eq:value_alignment}), the two
branches are
\begin{equation}
\begin{aligned}
O^{D}
&=
\operatorname{Softmax}
\left(
\frac{
Q^{\mathrm{cur}}
[K^{\mathrm{cur}};K^{\mathrm{ctx}};K^{D}]^{\top}
}{
\sqrt{d_h}
}
+
[B^{\mathrm{cur}};0;B^{D}]
\right)
[V^{\mathrm{cur}};V^{\mathrm{ctx}};\bar V^{D}],
\\
O^{S}
&=
\operatorname{Softmax}
\left(
\frac{
Q^{\mathrm{cur}}
[K^{\mathrm{cur}};K^{S}]^{\top}
}{
\sqrt{d_h}
}
+
[B^{\mathrm{cur}};B^{S}]
\right)
[V^{\mathrm{cur}};\bar V^{S}].
\end{aligned}
\label{eq:dual_branch_attention}
\end{equation}
Here, $B^{\mathrm{cur}}$ preserves native causal attention,
$B^D$ restricts dynamic attention according to both projected query
support and valid dynamic keys, and $B^S$ masks invalid static keys while being shared across all current queries. The two branches are then fused via normalized linear blending.

\textbf{Timestep-Aware Memory Routing.}
The two memory components contribute differently across the denoising trajectory.
At early, high-noise timesteps, static references provide spatial and appearance anchors for establishing coarse scene structure, while dynamic references supply historical entity evidence for temporal continuity.
As denoising progresses and the scene structure becomes established, continued static conditioning may over-constrain fine-grained visual refinement.
GROME therefore activates both static and dynamic branches during early denoising, while retaining only the dynamic branch at later timesteps.
This timestep-aware routing enables effective coarse-to-fine memory conditioning without modifying the pretrained backbone.
\begin{figure*}[t]
  \centering
  \includegraphics[width=\textwidth]{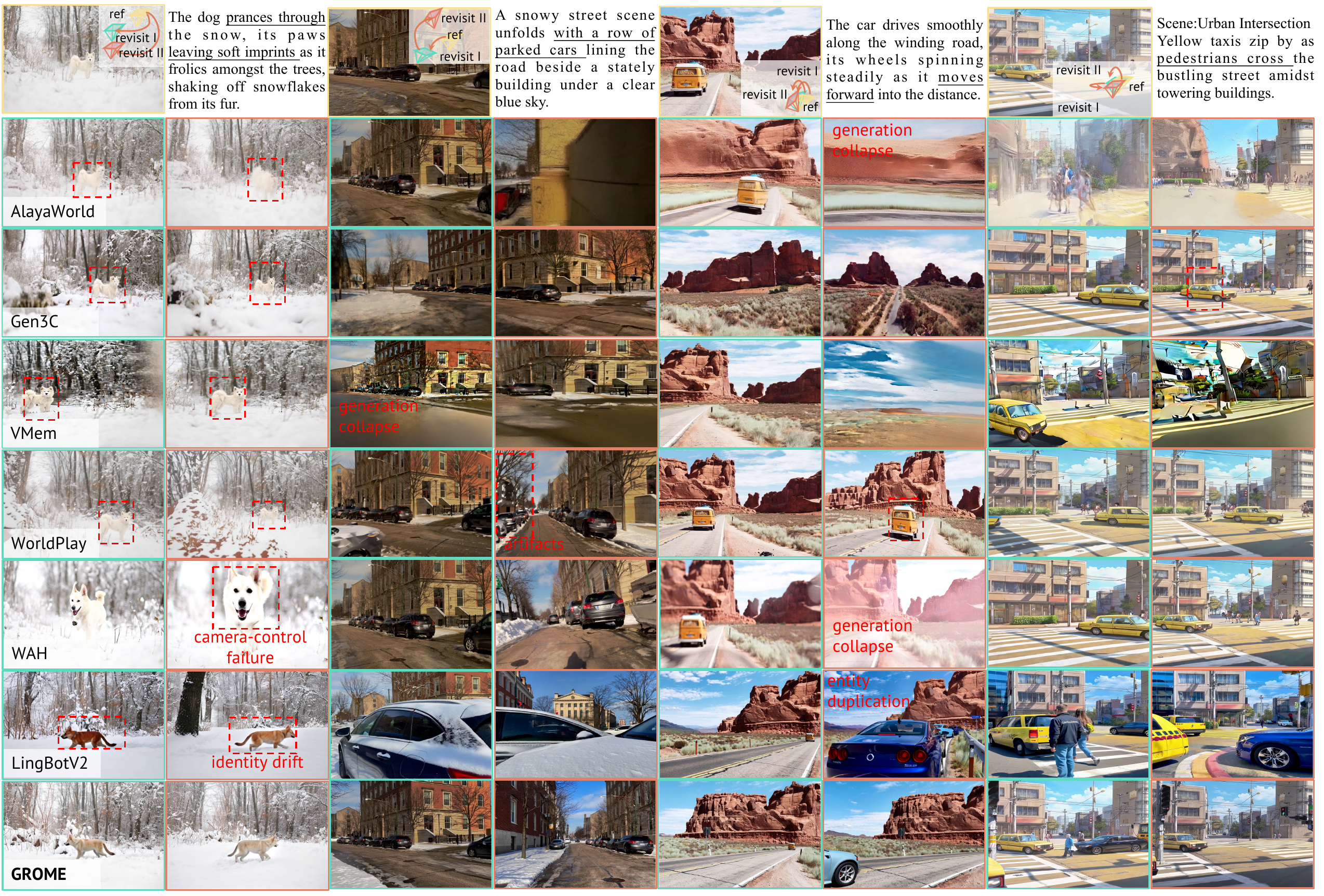}
  \caption{
    \textbf{Qualitative comparison.}
    GROME better preserves revisit scene structure and  entity consistency simultaneously across diverse revisit trajectories compared with existing world models.}
  \label{fig:compare}
\end{figure*}
\section{Experiments}

\subsection{Experimental Setup}
\label{sec:experimental_setup}

\textbf{Implementation Details.}
GROME operates as a training-free memory framework that enhances frozen world models for long-horizon consistency without any parameter updates.
Both GROME and its controlled variants build upon LingBotV2~\citep{gao2026infinite}.
Memory injection is deliberately scheduled across denoising: static scene references are injected solely during the first two denoising steps to establish global layout, whereas dynamic entity references remain active throughout denoising to ensure continuous temporal evolution.
All external baselines strictly follow their official implementations and default sampling protocols on identical evaluation subsets.
Complete architecture parameters and inference hyperparameters are detailed in Appendix~\ref{app:evaluation_protocol}.

\textbf{Evaluation Protocol and Metrics.} We compare GROME with mainstream baselines under a unified long-horizon revisit protocol from three complementary aspects. The benchmark contains 200 WorldScore scenes~\citep{duan2025worldscore}, including 100 static and 100 dynamic scenes, with 15--18\,s trajectories at 24 FPS.
Long-horizon consistency is evaluated at matched revisit poses using PSNR, SSIM, and LPIPS. To assess the quality of generated videos, including subject preservation and dynamic behavior, we report Entity Consistency, Motion Accuracy, CLIP Aesthetic, and Subjective Quality. We further evaluate whether memory conditioning preserves camera and semantic controllability using camera translation/rotation errors and Prompt Consistency. Dataset composition, trajectory construction, and metric implementations are detailed in Appendix~\ref{app:evaluation_protocol}.

\begin{table*}[t]
  \centering
  \caption{
  Quantitative comparison on static and dynamic long-horizon revisit trajectories.
  LingBotV2$^{\mathrm{RC}}$ denotes LingBotV2 with raw-context conditioning,
  while LingBotV2$^{\mathrm{TR}}$ denotes LingBotV2 with temporal retrieval.
  Values are reported as mean $\pm$ the half-width of the 95\% paired bootstrap confidence intervals. Best and second-best values are shown in \textbf{bold} and \underline{underlined}.
  }
  \label{tab:main_results}

  \small
  \setlength{\tabcolsep}{2pt}

  \resizebox{\textwidth}{!}{%
  \begin{tabular}{lccc|cc|cc|ccc}
  \toprule

  & \multicolumn{3}{c|}{\textbf{Revisit Consistency}}
  & \multicolumn{2}{c|}{\textbf{Visual Quality}}
  & \multicolumn{2}{c|}{\textbf{Dynamic Quality}}
  & \multicolumn{3}{c}{\textbf{Controllability}} \\

  \cmidrule(lr){2-4}
  \cmidrule(lr){5-6}
  \cmidrule(lr){7-8}
  \cmidrule(lr){9-11}

  \textbf{Model}
  & PSNR $\uparrow$
  & SSIM $\uparrow$
  & LPIPS $\downarrow$
  & CLIP Aesthetic $\uparrow$
  & Subjective Qual. $\uparrow$
  & Entity Cons. $\uparrow$
  & Motion Acc. $\uparrow$
  & Trans. Err. $\downarrow$
  & Rot. Err. ($^\circ$) $\downarrow$
  & Prompt Cons. $\uparrow$ \\

  \midrule
  \multicolumn{11}{l}{\textbf{Static}} \\
  \midrule

  Gen3C
  & \bestcell{\cival{21.725}{0.5582}}
  & \bestcell{\cival{0.749}{0.0203}}
  & \underline{\cival{0.173}{0.0140}}
  & \cival{5.334}{0.0874}
  & \cival{0.469}{0.0264}
  & \nacell
  & \nacell
  & \underline{\cival{0.420}{0.0513}}
  & \bestcell{\cival{3.878}{0.6494}}
  & \cival{0.258}{0.0093} \\

  VMem
  & \cival{15.784}{0.5323}
  & \cival{0.460}{0.0221}
  & \cival{0.367}{0.0132}
  & \cival{5.089}{0.0760}
  & \cival{0.338}{0.0230}
  & \nacell
  & \nacell
  & \cival{0.624}{0.4502}
  & \cival{7.187}{1.0210}
  & \cival{0.251}{0.0121} \\

  WAH
  & \cival{16.911}{0.6472}
  & \cival{0.540}{0.0320}
  & \cival{0.369}{0.0266}
  & \cival{5.116}{0.0739}
  & \cival{0.347}{0.0308}
  & \nacell
  & \nacell
  & \cival{2.141}{0.7641}
  & \cival{6.709}{1.3273}
  & \cival{0.231}{0.0104} \\

  WorldPlay
  & \cival{16.087}{0.4921}
  & \cival{0.465}{0.0236}
  & \cival{0.289}{0.0139}
  & \cival{5.265}{0.0869}
  & \cival{0.404}{0.0265}
  & \nacell
  & \nacell
  & \cival{0.471}{0.0504}
  & \cival{6.391}{0.9905}
  & \cival{0.260}{0.0090} \\

  AlayaWorld
  & \cival{15.367}{0.5238}
  & \cival{0.487}{0.0340}
  & \cival{0.443}{0.0272}
  & \cival{4.911}{0.0745}
  & \cival{0.263}{0.0247}
  & \nacell
  & \nacell
  & \cival{0.766}{0.2868}
  & \cival{7.829}{1.1486}
  & \cival{0.241}{0.0105} \\

  LingBotV2$^{\mathrm{RC}}$
  & \cival{11.400}{0.3338}
  & \cival{0.285}{0.0175}
  & \cival{0.590}{0.0152}
  & \underline{\cival{5.355}{0.0740}}
  & \underline{\cival{0.558}{0.0263}}
  & \nacell
  & \nacell
  & \cival{1.620}{0.4639}
  & \cival{12.156}{1.1314}
  & \underline{\cival{0.269}{0.0084}} \\

  LingBotV2$^{\mathrm{TR}}$
  & \cival{12.805}{0.2783}
  & \cival{0.291}{0.0180}
  & \cival{0.597}{0.0097}
  & \cival{5.278}{0.0617}
  & \cival{0.514}{0.0248}	
  & \nacell
  & \nacell
  & \cival{1.143}{0.2427}
  & \cival{15.698}{6.2615}
  & \cival{0.265}{0.0086} \\

  \textbf{GROME}
  & \underline{\cival{20.620}{0.4196}}
  & \underline{\cival{0.652}{0.0266}}
  & \bestcell{\cival{0.169}{0.0178}}
  & \bestcell{\cival{5.436}{0.0890}}
  & \bestcell{\cival{0.581}{0.0305}}
  & \nacell
  & \nacell
  & \bestcell{\cival{0.401}{0.0368}}
  & \underline{\cival{3.984}{0.8028}}
  & \bestcell{\cival{0.270}{0.0092}} \\

  \midrule
\multicolumn{11}{l}{\textbf{Dynamic}} \\
\midrule

Gen3C
& \cival{22.183}{1.3030}
& \cival{0.696}{0.0415}
& \cival{0.259}{0.0226}
& \cival{5.037}{0.1293}
& \cival{0.473}{0.0476}
& \underline{\cival{0.841}{0.0125}}
& \cival{-38.560}{25.0853}
& \bestcell{\cival{0.417}{0.0238}}
& \underline{\cival{0.835}{0.5012}}
& \cival{0.213}{0.0113} \\

VMem
& \cival{17.448}{1.6641}
& \cival{0.551}{0.0226}
& \cival{0.349}{0.0329}
& \cival{4.916}{0.2246}
& \cival{0.313}{0.0840}
& \cival{0.461}{0.0253}
& \cival{-2.324}{25.6885}
& \cival{1.242}{0.5574}
& \cival{9.336}{6.2733}
& \cival{0.203}{0.0312} \\

WAH
& \cival{23.423}{1.4130}
& \underline{\cival{0.703}{0.0143}}
& \cival{0.259}{0.0125}
& \cival{4.999}{0.1348}
& \cival{0.406}{0.0527}
& \cival{0.748}{0.0155}
& \cival{14.475}{3.5071}
& \cival{3.090}{3.1861}
& \cival{2.568}{2.1433}
& \cival{0.219}{0.0106} \\

WorldPlay
& \underline{\cival{23.870}{1.0117}}
& \cival{0.654}{0.0337}
& \underline{\cival{0.249}{0.0179}}
& \cival{5.120}{0.1481}
& \cival{0.492}{0.0541}
& \bestcell{\cival{0.910}{0.0113}}
& \cival{-2.883}{12.2510}
& \cival{0.695}{0.0106}
& \cival{0.963}{0.4839}
& \cival{0.214}{0.0104} \\

AlayaWorld
& \cival{22.479}{1.2827}
& \cival{0.656}{0.0377}
& \cival{0.332}{0.0265}
& \cival{4.778}{0.1282}
& \cival{0.321}{0.0543}
& \cival{0.764}{0.0240}
& \cival{-15.656}{6.0129}
& \cival{1.887}{1.6161}
& \cival{10.420}{7.3239}
& \cival{0.207}{0.0150} \\

LingBotV2$^{\mathrm{RC}}$
& \cival{19.291}{0.9170}
& \cival{0.638}{0.0320}
& \cival{0.403}{0.0284}
& \cival{5.199}{0.1052}
& \underline{\cival{0.557}{0.0358}}
& \cival{0.717}{0.0191}
& \underline{\cival{16.640}{16.1201}}
& \cival{2.160}{1.5836}
& \cival{4.561}{1.6069}
& \underline{\cival{0.227}{0.0114}} \\

LingBotV2$^{\mathrm{TR}}$
& \cival{21.243}{1.0685}
& \cival{0.700}{0.0257}
& \cival{0.312}{0.0290}
& \underline{\cival{5.259}{0.1069}}
& \cival{0.415}{0.0419}
& \cival{0.691}{0.0234}
& \cival{14.222}{12.2639}
& \cival{3.681}{2.3274}
& \cival{7.084}{1.8285}
& \cival{0.224}{0.0105} \\

\textbf{GROME}
& \bestcell{\cival{24.111}{0.9633}}
& \bestcell{\cival{0.717}{0.0437}}
& \bestcell{\cival{0.243}{0.0263}}
& \bestcell{\cival{5.274}{0.1047}}
& \bestcell{\cival{0.585}{0.0481}}
& \cival{0.769}{0.0274}
& \bestcell{\cival{16.840}{14.6834}}
& \underline{\cival{0.498}{0.1041}}
& \bestcell{\cival{0.759}{0.5400}}
& \bestcell{\cival{0.230}{0.0130}} \\

\bottomrule

  \end{tabular}
  }
\end{table*}

\subsection{Main Results and Analysis}
\label{sec:results}

\textbf{Quality Results.}
As illustrated in Fig.~\ref{fig:compare}, existing world models exhibit distinct trade-offs across long-horizon revisits, dynamic entity consistency, and trajectory controllability. While Gen3C and WorldPlay preserve plausible appearance consistency for dynamic subjects, their motion tends to freeze over extended horizons. Conversely, WAH sustains dynamic motion but struggles to strictly follow the intended camera trajectory, leading to degraded camera controllability during revisits. LingBotV2 demonstrates competitive camera tracking and dynamic motion; however, over extended rollouts, its dynamic entities suffer from notable drift, while the model frequently hallucinates spurious or duplicate objects (e.g., redundant cars), thereby degrading visual coherence and deviating from the prompt. In contrast, GROME effectively balances these competing objectives: it retains consistent scene geometry upon revisits, preserves persistent dynamic entities with continuous motion, and faithfully adheres to both the target camera trajectory and text prompt.

\textbf{Quantitative Results.}
Table~\ref{tab:main_results} evaluates \textbf{GROME} against representative world modeling baselines.
Notably, existing methods struggle with a dilemma between long-horizon revisit consistency and motion dynamics, whereas GROME offers a better joint trade-off between them. While Gen3C and WorldPlay achieve high entity consistency, they yield severely negative motion accuracy. This exposes a fundamental failure mode in coupled spatial representations: to enforce scene persistence, their coupled representations can collapse dynamic entities into static or frozen textures. In contrast, GROME achieves positive motion accuracy while preserving high entity consistency, demonstrating that factorizing historical observations into decoupled static and dynamic components prevents environmental persistence from penalizing dynamic motion.
To verify that these gains stem specifically from factorized memory rather than backbone capacity or conditioning budgets, we benchmark against LingBotV2$^{\mathrm{RC}}$ and LingBotV2$^{\mathrm{TR}}$ under the same architecture. On static scenes, naive temporal retrieval only marginally improves PSNR, as unconstrained temporal retrieval struggles to resolve geometric ambiguities across wide baselines. Conversely, GROME significantly elevates PSNR and markedly improves camera controllability. On dynamic scenes, GROME consistently outperforms retrieval baselines in rendering fidelity and revisit consistency. These controlled comparisons on LingBotV2 confirm the superiority of GROME, showing that long-horizon world modeling requires structured static-dynamic factorization rather than simple historical access.
\subsection{Ablation Study}
\label{sec:ablation}
\textbf{Static-Dynamic Factorization.}
Removing the static component substantially lowers PSNR and increases camera rotation error, showing that persistent scene evidence provides the geometric constraints required to recover previously observed views.
Removing the dynamic component instead reduces entity consistency while maintaining relatively low camera errors, indicating that scene geometry alone cannot preserve evolving entities over long rollouts.
These complementary effects show that GROME preserves static scene structure and dynamic entities through factorized but cooperative memory components over extended autoregressive generation horizons.
\begin{table*}[t]
  \centering
  \caption{
  Quantitative ablation on Static-Dynamic Factorization and Injection across 50 scenes.
  }
  \label{tab:ablation}
  \small
  \setlength{\tabcolsep}{2pt}
  \resizebox{\textwidth}{!}{%
  \begin{tabular}{lccc|cc|cc|ccc}
  \toprule
  & \multicolumn{3}{c|}{\textbf{Revisit Consistency}}
  & \multicolumn{2}{c|}{\textbf{Visual Quality}}
  & \multicolumn{2}{c|}{\textbf{Dynamic Quality}}
  & \multicolumn{3}{c}{\textbf{Controllability}} \\
  \cmidrule(lr){2-4}
  \cmidrule(lr){5-6}
  \cmidrule(lr){7-8}
  \cmidrule(lr){9-11}

  \textbf{Setting}
  & PSNR $\uparrow$
  & SSIM $\uparrow$
  & LPIPS $\downarrow$
  & CLIP Aesthetic $\uparrow$
  & Subjective Qual. $\uparrow$
  & Entity Cons. $\uparrow$
  & Motion Acc. $\uparrow$
  & Trans. Err. $\downarrow$
  & Rot. Err. ($^\circ$) $\downarrow$
  & Prompt Cons. $\uparrow$ \\
  \midrule

  LingBotV2$^{\mathrm{RC}}$
  & \cival{15.505}{3.1742}
  & \cival{0.456}{0.0418}
  & \cival{0.500}{0.0147}
  & \cival{5.247}{0.2950}
  & \underline{\cival{0.572}{0.0875}}
  & \cival{0.711}{0.0482}
  & \underline{\cival{16.333}{13.4337}}
  & \cival{1.903}{1.2061}
  & \cival{8.391}{1.2254}
  & \cival{\underline{0.247}}{0.0103} \\

  Joint Attention.
  & \cival{5.470}{0.8511}
  & \cival{0.203}{0.0115}
  & \cival{0.533}{0.0739}
  & \cival{4.417}{0.0275}
  & \cival{0.153}{0.0430}
  & \cival{0.641}{0.0170}
  & \cival{10.987}{17.7273}
  & \cival{2.573}{0.1557}
  & \cival{8.834}{3.4000}
  & \cival{0.224}{0.0142} \\

  w/o Static
  & \cival{18.512}{2.9186}
  & \cival{0.609}{0.0471}
  & \cival{0.323}{0.0152}
  & \underline{\cival{5.309}{0.3987}}
  & \cival{0.566}{0.1199}
  & \underline{\cival{0.735}{0.0160}}
  & \cival{15.709}{12.6554}
  & \cival{1.586}{0.0388}
  & \cival{7.252}{0.4953}
  & \cival{\underline{0.247}}{0.0111}\\

  w/o Dynamic
  & \underline{\cival{19.542}{3.2823}}
  & \underline{\cival{0.640}{0.0354}}
  & \underline{\cival{0.270}{0.0104}}
  & \cival{5.231}{0.3214}
  & \cival{0.550}{0.0464}
  & \cival{0.609}{0.0216}
  & \cival{15.430}{11.5595}
  & \cival{0.450}{0.1339}
  & \cival{2.867}{3.5811}
  & \cival{0.244}{0.0098} \\

  w/o Ref. Adapt
  & \cival{6.344}{2.7415}
  & \cival{0.233}{0.0446}
  & \cival{0.475}{0.0138}
  & \cival{4.524}{0.3682}
  & \cival{0.208}{0.1082}
  & \cival{0.682}{0.0120}
  & \cival{14.135}{12.5521}
  & \cival{0.468}{0.0339}
  & \cival{2.695}{0.4918}
  & \cival{0.242}{0.0124} \\

  Static at all Steps
  & \cival{16.003}{3.2694}
  & \cival{0.576}{0.1112}
  & \cival{0.317}{0.0133}
  & \cival{5.066}{0.0344}
  & \cival{0.515}{0.1162}
  & \cival{0.685}{0.0296}
  & \cival{14.507}{13.2834}
  & \underline{\cival{0.413}{0.0587}}
  & \underline{\cival{2.645}{3.5399}}
  & \cival{0.245}{0.0107} \\

  \midrule

  \textbf{Full GROME}
  & \bestcell{\cival{23.838}{2.5864}}
  & \bestcell{\cival{0.743}{0.0382}}
  & \bestcell{\cival{0.189}{0.0119}}
  & \bestcell{\cival{5.352}{0.1216}}
  & \bestcell{\cival{0.598}{0.0412}}
  & \bestcell{\cival{0.749}{0.0148}}
  & \bestcell{\cival{16.872}{12.3846}}
  & \bestcell{\cival{0.363}{0.0315}}
  & \bestcell{\cival{2.320}{0.4627}}
  & \bestcell{\cival{0.251}{0.0113}} \\

  \bottomrule
  \end{tabular}
  }
\end{table*}

\begin{figure*}[t]
  \centering
  \includegraphics[width=0.95\linewidth]{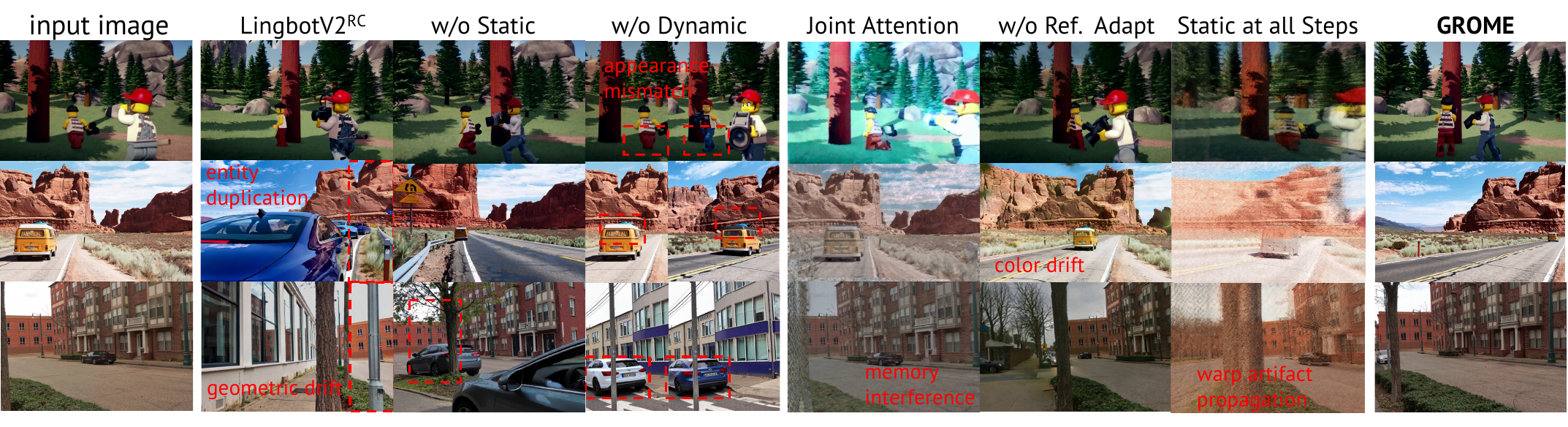}
  \caption{
    \textbf{Qualitative ablation of GROME.}
    Static-dynamic factorization and three injection strategies
    demonstrate their complementary roles in maintaining static scene and evolving entity states.
    }
\label{fig:ablation}
\end{figure*}

\textbf{Training-Free Memory Injection.}
Na\"ively injecting static and dynamic references through a shared attention pathway yields substantially lower PSNR than the full model, showing that additional historical evidence alone is insufficient.
Removing Reference Value Adaptation further degrades PSNR and causes severe appearance mismatch between projected references and native context, which can accumulate over autoregressive rollout and destabilize generation (Figure~\ref{fig:ablation}).
Likewise, injecting static references throughout denoising lowers PSNR, as geometric and warping errors persist into refinement; Timestep-Aware Memory Routing restricts static guidance to coarse structure formation. Together, these results demonstrate the complementary roles of these strategies.
\begin{wrapfigure}{r}{0.5\linewidth}
  \vspace{-10pt}
  \centering
  \includegraphics[width=1\linewidth]{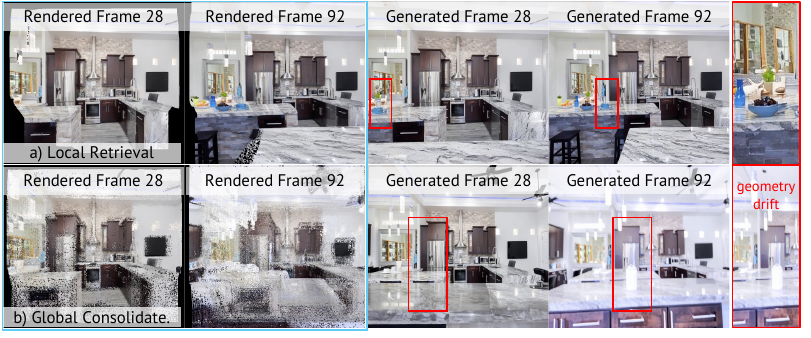}
  \caption{Local retrieval vs. Global consolidation.}
  \label{fig:local_global}
  \vspace{-10pt}
\end{wrapfigure}
\textbf{Local Retrieval vs. Global Consolidation.}
As shown in Fig.~\ref{fig:local_global}, repeatedly consolidating observations into a global cache accumulates geometric errors, producing visible drift in both rendered references and generated frames. In contrast, local retrieval preserves the geometry of individual observations and composes them only when needed, yielding more stable references and subsequent generation over rollouts.

\subsection{Applications}
\label{sec:applications}
\begin{wrapfigure}{r}{0.4\linewidth}
  \vspace{-20pt}
  \centering
  \includegraphics[width=1\linewidth]{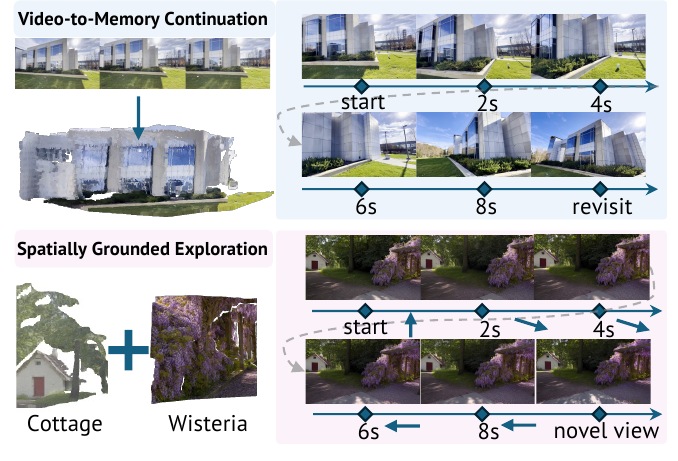}
  \caption{Applications of GROME.}
  \label{fig:applications}
  \vspace{-40pt}
\end{wrapfigure}

Benefiting from its persistent static-dynamic memory and long-horizon consistency, GROME enables several downstream applications, as shown in Fig.~\ref{fig:applications}. First, it can initialize memory directly from an input video and continue generation along a user-specified trajectory, allowing the rollout to explore beyond the observed region and later revisit previously observed content. Moreover, the explicit world-space memory provides an editable spatial interface, enabling scene elements to be recomposed before generation and subsequently explored along novel trajectories while preserving the imposed spatial layout.

\section{Conclusion}
GROME is a memory framework that grounds factorized memory into frozen video models through distinct retrieval and training-free injection strategies. Experimental results demonstrate that GROME significantly boosting revisit consistency over long rollouts. Moreover, GROME supports video-to-memory continuation and spatial scene editing, highlighting how structured explicit memory empowers models to remember, revisit, and evolve coherent worlds over extended horizons.

\clearpage
\section*{Ethics Statement}

This work follows the ICLR Code of Ethics. Our study concerns memory mechanisms
for video world models, evaluated on static and dynamic trajectories based on scenes in WorldScore~\citeyearpar{duan2025worldscore}.
Generated videos are synthetic simulations and should not be presented as
verified recordings of real events. GROME may inherit biases and limitations
from its frozen backbone and auxiliary models. Use and redistribution of
benchmark data, frozen models, and real-world video inputs remain subject
to their original licenses and applicable privacy requirements.

\section*{Reproducibility Statement}

GROME operates at inference time on a frozen LingBotV2 backbone and requires
no additional training. Section~\ref{sec:method} describes the static-dynamic memory factorization, local memory retrieval and training-free injection.
Appendix~\ref{app:memory} details depth alignment,
static-dynamic decomposition, and distinct memory retrieval policies. Algorithm~\ref{alg:GROME} summarizes
the inference procedure. Section~\ref{sec:experimental_setup} specifies the
WorldScore evaluation subset, revisit protocols, and reported metrics to
support replication of the evaluation.

\section*{AI Use Statement}

We used large language models to polish the manuscript, including rephrasing
sentences, correcting grammar, and improving readability and flow. The authors
reviewed and edited the AI-assisted text to ensure that it accurately reflects
the methods, experiments, and results, and take full responsibility for the
content of the manuscript.

\bibliographystyle{iclr2027_conference}
\bibliography{iclr2027_conference}
\clearpage
\appendix

\setcounter{tocdepth}{2}
\tableofcontents
\section{Overview of Supplementary Material}
This supplementary material provides implementation details, extended analyses, and additional experimental results to complement the main paper.

Specifically, we elaborate on:
\begin{itemize}
\item \textbf{Methodological Details} (Sec.~\ref{app:memory}): static-dynamic memory factorization, local memory retrieval, and training-free injection.
\item \textbf{Experimental Setup} (Sec.~\ref{app:evaluation_protocol}):
benchmark curation, baselines, and evaluation protocols.
\item \textbf{Extended Results} (Secs.~\ref{app:efficiency},
\ref{app:complex_memory}, and \ref{app:long_horizon_comparison}): runtime efficiency, complex-memory revisits, long-horizon comparisons, and additional qualitative analyses.
\item \textbf{Downstream Applications \& Limitations} (Secs.~\ref{app:applications} and \ref{app:limitations}): real-world video continuation, spatially guided exploration, and failure cases.
\end{itemize}
In addition, interactive video demonstrations are provided in the supplementary package. Readers are invited to unzip the folder and open \texttt{index.html} in a web browser to view animated examples of long-horizon revisit consistency, dynamic entity evolution, and interactive exploration showcased in the paper.

\section{Additional method details}
\label{app:method_details}
\begin{algorithm}[H]
  \caption{GROME inference overview.}
  \label{alg:GROME}
  \begin{algorithmic}[1]
  \REQUIRE Frozen world model $\mathcal{F}_{\theta}$,
  initial frame $X_0$,
  action sequence $\{a_t\}$
  \ENSURE Long-horizon generated rollout
  \STATE Initialize active context $\mathcal{R}^{\mathrm{ctx}}$ and
  factorized memory $\mathcal{M}=(\mathcal{M}^{S},\mathcal{M}^{D})$
  \FOR{each generation step $t$}
      \STATE Project $\mathcal M_{t-1}$ along the target camera trajectory $q_t$, combining static evidence and resolving overlapping dynamic evidence by recency
      \STATE Encode the projected historical references and condition the frozen model through separate attention branches
      \STATE Generate chunk $X_t$
      \STATE Separate $X_t$ into static scene content and dynamic entities and store their local pointmaps in $\mathcal M_t$
      \STATE Update active context with the newest generated frames
  \ENDFOR
  \RETURN generated rollout
  \end{algorithmic}
\end{algorithm}

At generation step $t$, the model reads the factorized memory
$\mathcal M_{t-1}$, projects it along the target camera trajectory $q_t$,
and uses the resulting references to generate chunk $X_t$. After generation,
the frames of $X_t$ are processed and added to $\mathcal M_t$ for later steps.

\subsection{Static-Dynamic Memory Factorization}
\label{app:memory}
\label{app:memory_construction}

This section expands the memory construction in Sec.~\ref{sec:method}. GROME
processes generated frames, aligns their estimated depth to the static component
of factorized memory, and partitions the resulting spatial content into static
scene regions and dynamic entity regions. 

\subsubsection{Depth alignment across chunks}
\label{app:depth_alignment}

For each generated chunk $X_t$, GROME predicts a metric monocular depth map $\hat D_{t,f}$ for each frame $X_{t,f}$ using DA3~\citep{lin2025depth} with overlapping chunk-wise prediction. Let $\mathcal F_t$ be the frame indices in the current geometry update, and let $\mathcal O_t\subseteq\mathcal F_t$ be those shared with the preceding window. These shared frames provide correspondence for depth calibration, while newly generated frames extend the reconstructed region. Each frame is aligned independently to the existing static component whenever it has sufficient valid projected overlap.
Although metric depth provides a common depth scale, residual scale and offset inconsistencies can remain across autoregressive chunks and lead to misalignment when their geometry is later reused.

To align each frame with previously stored static evidence, GROME
projects the static component $\mathcal M_{t-1}^S$ into that frame's camera view
to obtain a reference depth
\begin{equation}
D_{t,f}^{\mathrm{ref}}
=
\Pi_D\!\left(
\mathcal M_{t-1}^S;
C_{t,f}
\right),
\qquad
 f\in\mathcal F_t,
\label{eq:memory_reference_depth}
\end{equation}
where $C_{t,f}$ denotes the camera parameters of frame $f$ in the current geometry update. Each projected reference depth is accompanied by its valid visibility mask $M_{t,f}^{\mathrm{ref}}$. The overlapping frames $f\in\mathcal O_t$ anchor consecutive DA3 windows; the same frame-wise fit applies to a newly generated frame when its projected static memory provides sufficient overlap.

\textbf{Initialization.}
For the first generated chunk, no previous static component is available and no cross-frame alignment is performed. We therefore set $D_{t,f}=\hat D_{t,f}$ for its frames and use them to initialize the memory. Subsequent depth predictions are calibrated against the static content established by previously processed frames.

Depth alignment is restricted to static regions shared by the current
generated frame and the projected memory:
\begin{equation}
\Omega_{t,f}^{\mathrm{align}}
=
M_{t,f}^{\mathrm{ref}}
\cap
M_{t,f}^{\mathrm{static}}.
\label{eq:depth_alignment_region}
\end{equation}
Here $M_{t,f}^{\mathrm{static}}$ excludes regions identified as dynamic by the
SAM3~\citep{carion2026sam}/CoTracker~\citep{karaev2024cotracker} motion-aware
segmentation.
Frame $f$ is aligned only when
$|\Omega_{t,f}^{\mathrm{align}}|\geq256$ pixels.
Otherwise, depth alignment is skipped for that frame.

\textbf{Affine Inverse-Depth Alignment.}
We perform alignment in inverse-depth space.
Let
$\hat q_{t,f}(u)=1/\hat D_{t,f}(u)$ and
$q_{t,f}^{\mathrm{ref}}(u)=1/D_{t,f}^{\mathrm{ref}}(u)$.
Before fitting, both inverse-depth distributions are trimmed to their
$[0.10,0.90]$ quantile ranges over $\Omega_{t,f}^{\mathrm{align}}$.
On the remaining pixels $\Omega_{t,f}^{\mathrm{fit}}$, we estimate a separate affine correction for each frame:
\begin{equation}
(\alpha_{t,f},\beta_{t,f})
=
\arg\min_{\alpha,\beta}
\sum_{u\in\Omega_{t,f}^{\mathrm{fit}}}
\left(
\alpha\,\hat q_{t,f}(u)+\beta-q_{t,f}^{\mathrm{ref}}(u)
\right)^2.
\label{eq:depth_alignment_opt}
\end{equation}
The pair $(\alpha_{t,f},\beta_{t,f})$ is fitted independently for each $f$; no affine parameters are shared across frames in the update.
We use ordinary least squares without an additional robust loss or RANSAC;
outliers are removed only by the quantile trimming above.
If the source inverse-depth variance is below $10^{-8}$, we set the affine
bias to zero and estimate the scale from the median inverse-depth ratio.

The aligned depth is then
\begin{equation}
D_{t,f}(u)
=
\frac{1}{\alpha_{t,f}/\hat D_{t,f}(u)+\beta_{t,f}}.
\label{eq:aligned_depth}
\end{equation}
Pixels whose corrected inverse depth is non-finite or non-positive are
excluded from the subsequent memory construction.
This alignment places each newly generated frame in a depth gauge
consistent with the static geometry projected from previously stored memory.

\subsubsection{Static-Dynamic Decomposition}
\label{app:factorized_memory}
To construct decoupled world memories, we first separate transient dynamic entities from persistent static environments by compensating for ego-motion. Specifically, given video frames, we apply SAM3~\citep{carion2026sam} to segment instance masks and leverage CoTracker~\citep{karaev2024cotracker} to track dense point trajectories. Since camera motion introduces apparent displacement to static regions, we estimate the global camera-induced motion from background point trajectories and compute the relative motion residuals of candidate instances; an entity is classified as dynamic only when its trajectory significantly diverges from the estimated background field. Guided by the resulting static and dynamic masks, the local pointmap of each frame is factorized into static and dynamic sub-clouds, $P_{i,f}^S$ and $P_{i,f}^D$. Both are stored as local pointmaps registered with their corresponding camera poses, while dynamic pointmaps are additionally indexed with source timestamps to facilitate temporal-recency-based retrieval during future generation.
\subsubsection{Local Memory Retrieval}
\label{app:memory_projection}

\textbf{Static Retrieval and Consolidation.}
Rather than maintaining an accumulated global point cloud that is prone to drift, the static component preserves individual local pointmaps $P_{i,f}^S$ in their original camera coordinate frames alongside camera-to-world poses $T_{i,f}$, as defined in Eq.~\ref{eq:factorized_memory}. Retaining historical observations as decoupled local evidence avoids the artifacts of repeated point fusion across independently estimated geometries.
Conditioned on a target camera trajectory, GROME selects a set of complementary source maps via the max-coverage criterion of AnchorWeave~\citep{wang2026anchorweave}. These selected maps are synthesized into target-view references and encoded by the frozen VAE into at most $N_S$ latent slots, thereby decoupling the source observation count from the model's fixed token budget.

\textbf{Dynamic States Selection.}
For dynamic memory, historical entities across different timestamps may project onto overlapping regions in the target view, inducing spatial-temporal ambiguity. GROME resolves these conflicts through a region-based recency policy: candidate bounding boxes of projected entities that overlap are clustered, and only the observation with the most recent source timestamp is retained, actively suppressing stale historical duplicates. The disambiguated static and dynamic pointmaps are subsequently rasterized via standard depth-based z-buffering. For positional encoding, dynamic references adopt target-view 2D coordinates for spatial RoPE while preserving their original source timestamps for temporal RoPE, delivering geometrically aligned yet temporally faithful conditioning.

\subsection{Generation configuration}
\label{app:generation_config}
The controlled LingBotV2 comparisons generate at 24 FPS and follow the input
image resolution; the default resolution bucket is $480\times832$.
Autoregressive generation uses four-latent chunks and four diffusion steps
without classifier-free guidance. 
For GROME, the static-reference branch uses a fixed branch weight of
\begin{equation}
\alpha_S = 0.5,
\end{equation}
and is activated only during the first two denoising steps of the four-step
sampling process. The dynamic-reference branch remains active throughout all four
denoising steps. This schedule and branch weight are fixed across all
evaluation scenes and trajectories.
For Reference Value Adaptation, we use fixed alignment strengths
\begin{equation}
\lambda_S = 0.2,
\qquad
\lambda_D = 0.2,
\end{equation}
for the static and dynamic reference streams, respectively. These values are kept unchanged across all evaluation scenes and trajectories.
We use the same random seed for all controlled LingBotV2 comparisons on a given trajectory; external baselines use their official sampling configurations.

\textbf{Bootstrap Confidence Intervals.}
We estimate uncertainty by bootstrapping case- or video-level metric values,
rather than individual frames. 
For a metric with valid values $\mathcal X=\{x_i\}_{i=1}^{N}$, each bootstrap replicate samples $N$ evaluation units with replacement and computes the corresponding mean. For all methods, the same resampled indices are used across methods within each replicate to preserve the paired evaluation structure.
We use $10{,}000$ bootstrap replicates and compute 95\% confidence intervals with the percentile method, taking the $2.5$th and $97.5$th percentiles of the bootstrap mean distribution as the lower and upper bounds.
All bootstrap experiments use a fixed random seed of $12345$. For compact presentation, result tables report the mean together with the half-width $(U-L)/2$ of the corresponding 95\% percentile interval.

\section{Evaluation Protocol}
\label{app:evaluation_protocol}

\subsection{Baseline Models}
\label{app:baseline_models}

We compare GROME with representative long-horizon world-modeling approaches
that differ in how historical information is retained and reused.
Table~\ref{tab:memory_taxonomy} organizes these methods by their primary way of
accessing generated history. ``Frame based'' denotes methods that retain
temporally organized frames, while ``pose indexed'' denotes methods that store
content with camera geometry in a shared coordinate system. This organization
describes the main access pattern rather than a complete ranking of each
method's capabilities.

\begin{table}[t]
  \centering
  \caption{
    \textbf{Comparison of historical memory access patterns in long-horizon world models.}
    \emph{Implicit memory} methods rely on temporally indexed 2D frames, whereas \emph{explicit memory} methods maintain global 3D reconstructions. 
    In contrast, \textbf{GROME} introduces a unified, factorized memory: both static and dynamic components are grounded as local pointmaps, with the dynamic stream retaining source timestamps for temporal indexing.
  }
  \label{tab:memory_taxonomy}
  \small
  \setlength{\tabcolsep}{5pt}
  \begin{tabular}{lll}
    \toprule
    Method
    & \makecell{Static\\Scene}
    & \makecell{Moving\\Regions} \\
    \midrule
    VMem~\citep{li2025vmem}
    & \temporal
    & \temporal \\

    WAH~\citep{Wang2026WarpasHistoryGC}
    & \temporal
    & \temporal \\

    AlayaWorld~\citep{team2026alayaworld}
    & \spatial
    & \temporal \\

    Gen3C~\citep{ren2025gen3c}
    & \spatial
    & \spatial \\

    WorldPlay~\citep{sun2025worldplay}
    & \temporal
    & \temporal \\

    LingBotV2$^{\mathrm{RC}}$
    & \temporal
    & \temporal \\

    LingBotV2$^{\mathrm{TR}}$
    & \temporal
    & \temporal \\


    \textbf{GROME}
    & \memmark{spatialorange}\textbf{Spatial}
    & \memmark{spatiotemporalpink}\textbf{SpatioTemporal} \\
    \bottomrule
  \end{tabular}
\end{table}

\textbf{LingBotV2 Backbone.}
All controlled comparisons in our main experiments use the same pretrained
LingBotV2~\citep{gao2026infinite} backbone with frozen parameters.
The raw LingBotV2 baseline uses its built-in history-conditioning pathway
without external memory or retrieval. GROME and the temporal-retrieval
baseline use the same total conditioning budget; GROME additionally applies
the two attention branches described in Sec.~\ref{sec:method-registration}. The backbone parameters, sampling configuration, camera trajectory, and random seed are otherwise kept unchanged.

\textbf{Matched History-Conditioning Budget.}
All controlled LingBotV2 variants use the same total history-conditioning
capacity. The raw baseline uses recent context, temporal retrieval replaces part
of that context with FOV-overlap selected history, and GROME allocates the same
capacity across recent context and projected static and dynamic references. In
our experiments, the GROME allocation is 12 recent-context slots, 4 dynamic
reference slots, and 4 static reference slots, for a total of 20 latent slots.

\textbf{Temporal Retrieval Baseline.}
To isolate structured world memory from direct frame retrieval, the temporal
retrieval baseline ranks previously generated chunks by field-of-view overlap
with the target trajectory, following the co-visibility criterion of
Context-as-Memory~\citep{yu2025context}. Retrieved history replaces an equal
amount of recent context, so the raw baseline, temporal retrieval, and GROME
share the same total conditioning capacity and all other LingBotV2 settings.

\textbf{Other World-Model Baselines.}
For comparisons with independently developed world models
(VMem, WAH, WorldPlay, AlayaWorld, and Gen3C), we use their official
implementations and evaluation protocols whenever available.
These external baselines retain their official model-specific sampling and
inference configurations; only the WorldScore evaluation subset and prescribed
camera trajectories are shared across methods. By contrast, the controlled
LingBotV2 comparisons share resolution, FPS, random seed, sampling
configuration, camera trajectory, and history-conditioning budget.
These methods are included to compare different long-horizon memory
representations rather than as controlled architectural ablations of
LingBotV2. Controlled claims about history conditioning are therefore based
on the matched LingBotV2, temporal-retrieval, and GROME settings described
above.

\subsection{Evaluation Set Construction}
\label{app:benchmark_construction}

\textbf{Source Scenes.}
Our evaluation is constructed from 200 source scenes in
WorldScore~\citep{duan2025worldscore}, including 100 static and 100 dynamic
scenes. We retain the original scene content and prompts, while replacing
the original WorldScore generation protocol with our long-horizon
exploration-and-revisit trajectories.

\begin{figure}[H]
  \centering
  \includegraphics[width=\linewidth]{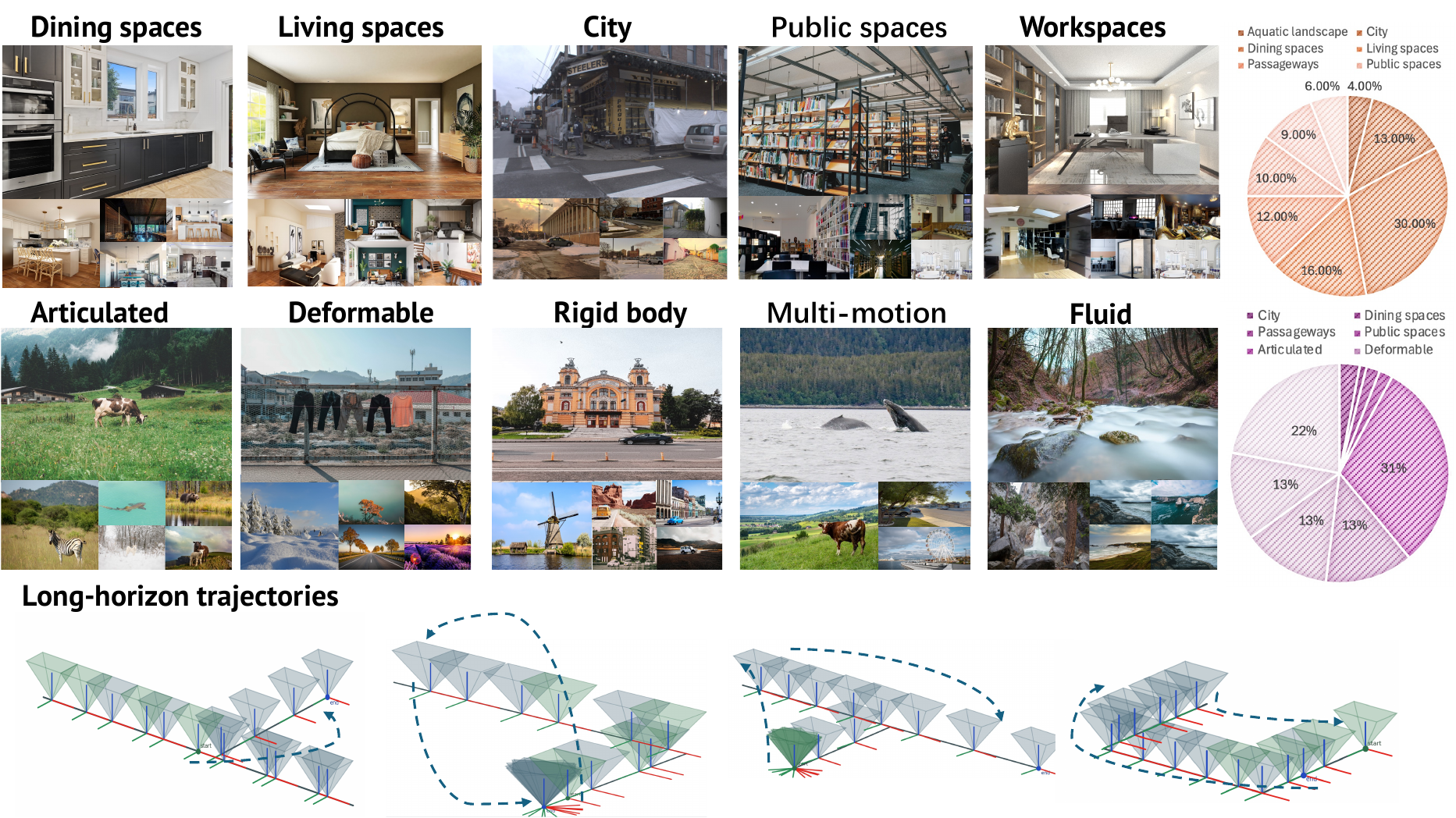}
  \caption{Evaluation set and trajectory design. The top panels show
  representative static and dynamic scene categories and their proportions.
  The bottom panels illustrate exploration-and-revisit camera trajectories;
  dashed arrows mark returns to previously observed regions.}
  \label{fig:eval_set_distribution}
\end{figure}

The subset is balanced across the scene categories and visual domains of
WorldScore. Figure~\ref{fig:eval_set_distribution} summarizes its composition.
The selected WorldScore sample IDs are listed in the anonymized supplementary configuration submitted with the paper.

\textbf{Difference From the Original WorldScore Protocol.}
WorldScore provides the scene content, prompts, and category annotations
used in our evaluation. We do not use its original generation trajectories.
The original benchmark evaluates static world generation using prescribed
scene transitions and evaluates dynamic generation with a fixed camera,
whereas our goal is to measure long-horizon memory during interactive
exploration. We therefore assign each selected scene a new camera trajectory
that first explores the environment and later revisits previously observed
regions. The same scene, prompt, and prescribed trajectory are used for all
compared methods.

\subsection{Long-horizon Trajectory Design}
\label{app:trajectory_design}

Each evaluation trajectory is represented by a sequence of prescribed camera
poses as shown in Figure~\ref{fig:eval_set_distribution},
\begin{equation}
\mathcal T
=
\{C_f\}_{f=1}^{F},
\qquad
C_f=(K_f,T_f),
\end{equation}
where $K_f$ and $T_f$ denote the camera intrinsics and extrinsics at frame
$f$. All videos are generated at 24 FPS, and each trajectory spans
15--18\,s, corresponding to 360--432 frames.

Each trajectory has an exploration phase followed by a revisit phase:
\begin{equation}
\mathcal T
=
\mathcal T^{\mathrm{explore}}
\cup
\mathcal T^{\mathrm{revisit}},
\end{equation}
where $\mathcal T^{\mathrm{explore}}$ moves the camera through previously
unobserved regions and $\mathcal T^{\mathrm{revisit}}$ later returns to camera
poses visited during the exploration phase. The revisit is placed sufficiently
late such that its corresponding reference frames have left the
backbone's effective context.

\subsection{Evaluation Metrics}
\label{app:evaluation_metrics}

Unless otherwise stated, we report raw component metrics rather than the normalized 0--100 scores derived from WorldScore's empirical bounds. This is particularly crucial for Motion Accuracy, whose unnormalized value is not bounded below by zero. Specifically, we categorize the evaluation protocol into three primary dimensions:

\textbf{Revisit Consistency.}
To assess long-horizon scene and instance persistence, we evaluate:
(1)~\emph{Static-Region Revisit Consistency}: We measure environmental persistence using PSNR, SSIM~\citep{wang2004image}, and LPIPS~\citep{zhang2018unreasonable} computed exclusively over jointly valid static regions. Dynamic pixels in either the reference or revisit frames are explicitly masked out to isolate background geometry from moving objects.
(2)~\emph{Entity Consistency}: We adapt the VBench Subject Consistency metric~\citep{huang2024vbench} to evaluate moving entities, computing feature-level appearance consistency strictly within SAM3-segmented dynamic instances rather than across the full frame.

\textbf{Visual and Dynamic Quality.}
To verify general generation fidelity and physical motion plausibility without memory degradation, we consider:
(1)~\emph{Visual Quality}: We evaluate perceived visual appeal and general realism using CLIP Aesthetic from VBench~\citep{huang2024vbench} and Subjective Quality from WorldScore~\citep{duan2025worldscore}.
(2)~\emph{Dynamic Quality}: We report the raw Motion Accuracy from WorldScore~\citep{duan2025worldscore} to quantify whether motion is properly concentrated within target regions. This metric penalizes spurious background movement and can yield negative values if out-of-region motion surpasses in-region dynamics.

\textbf{Controllability.}
To ensure the generative model faithfully adheres to user instructions and target trajectories, we measure:
(1)~\emph{Camera Pose Error}: We compute camera translation and rotation errors following the WorldScore~\citep{duan2025worldscore} protocol to evaluate 3D camera trajectory adherence.
(2)~\emph{Prompt Consistency}: We adopt the Overall Consistency metric from VBench~\citep{huang2024vbench} to assess semantic alignment with input text prompts across generated rollouts.

\section{Efficiency and Runtime Analysis}
\label{app:efficiency}

\textbf{Profiling Setup.}
GROME is training-free and introduces no additional optimization or
fine-tuning cost. All additional computation is incurred at inference time.

We profile a representative long-horizon dynamic rollout under the default
configuration on three NVIDIA A800 GPUs, where LingBotV2 is distributed over
two GPUs and the geometry pipeline runs on a separate GPU. The online geometry
estimator maintains a 93-frame sliding window with 77-frame overlap, while
each subsequent generation chunk contributes 16 new frames.

\begin{figure*}[t]
  \centering

  \begin{minipage}[t]{0.40\textwidth}
      \vspace{0pt}
      \centering
      \captionof{table}{
      DA3 inference-time scaling with input-frame count.
      The default 93-frame operating point is highlighted in light green.
      The 93-frame result reports mean and standard deviation over 51 updates.
      }
      \label{tab:da3_frame_scaling}
      \vspace{0.3em}
      \setlength{\tabcolsep}{6pt}
      \resizebox{\linewidth}{!}{%
      \begin{tabular}{ccc}
          \toprule
          \textbf{Input Frames}
          & \textbf{Samples}
          & \textbf{Time (s)} \\
          \midrule
          1  & 1  & 1.060 \\
          13 & 1  & 4.104 \\
          29 & 1  & 9.622 \\
          45 & 1  & 17.003 \\
          61 & 1  & 27.175 \\
          77 & 1  & 39.167 \\
          \profilecell{93}
          & \profilecell{51}
          & \profilecell{$52.437 \pm 0.994$} \\

          \bottomrule
      \end{tabular}
      }

  \end{minipage}
  \hfill
  \begin{minipage}[t]{0.45\textwidth}
      \vspace{0pt}
      \centering

      \includegraphics[width=0.8\linewidth]
      {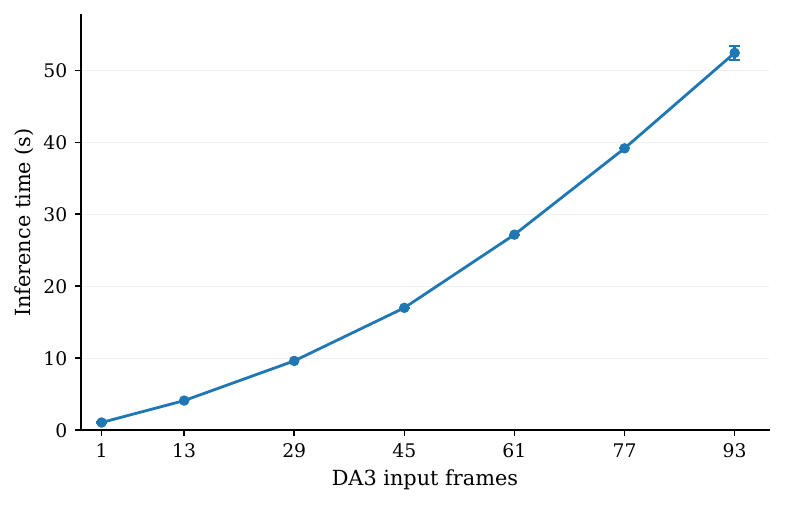}

      \captionof{figure}{
      \textbf{DA3 inference scaling with input-frame count.}
      Geometry-estimation cost increases with the active input-window size during initialization. GROME subsequently maintains a fixed sliding window instead of expanding the geometry input with rollout length.
      }
      \label{fig:da3_frame_scaling}

  \end{minipage}

\end{figure*}

\textbf{Scaling With Geometry Input Size.}
We first examine the computational cost of the geometry estimator itself.
Table~\ref{tab:da3_frame_scaling} reports pure DA3 inference time as its
active input window grows during initialization. Increasing the input from 1 to 93 frames raises the inference time from 1.060\,s to
$52.437\pm0.994$\,s. Once the 93-frame operating window is reached, its size remains fixed for subsequent updates.

\begin{figure*}[t]
  \centering

  \begin{minipage}[t]{0.49\textwidth}
      \centering
      \includegraphics[width=\linewidth]
      {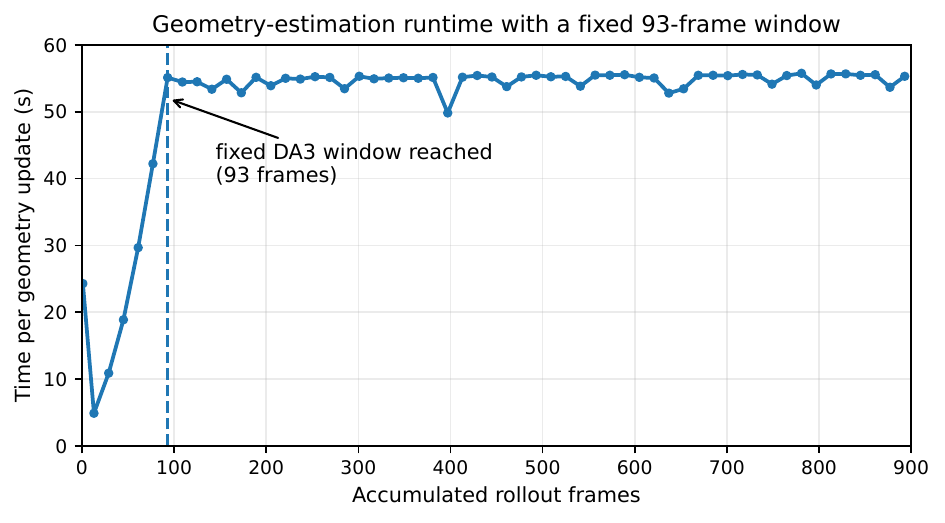}
      \vspace{-0.5em}
      \centerline{\small (a) Geometry-estimation runtime}
  \end{minipage}
  \hfill
  \begin{minipage}[t]{0.49\textwidth}
      \centering
      \includegraphics[width=\linewidth]
      {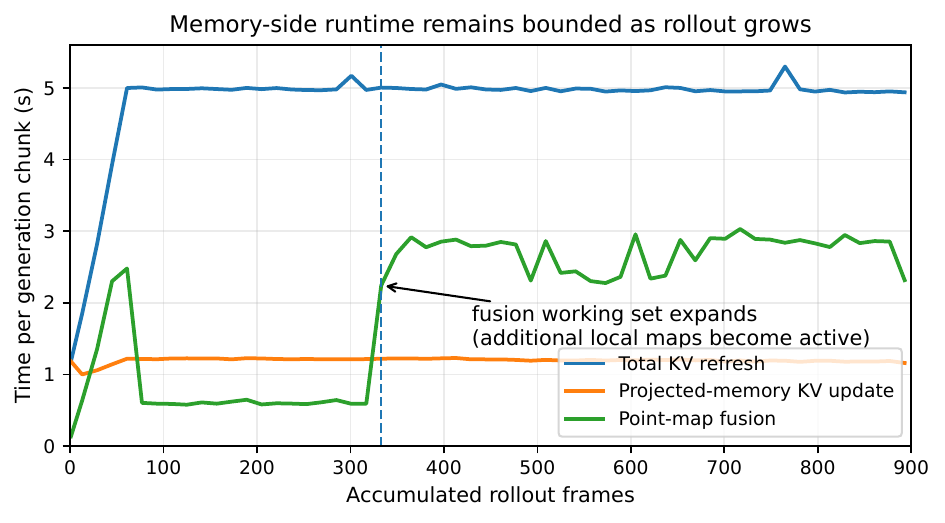}
      \vspace{-0.5em}
      \centerline{\small (b) Memory-side runtime}
  \end{minipage}
  \caption{
  \textbf{Runtime scaling with rollout length.}
  (a) During initialization, the DA3 input window grows to a fixed 93-frame window; subsequent geometry updates have approximately constant cost. (b) Reference K/V computation and total history-token preparation remain nearly constant with rollout length. The step increase in pointmap fusion around 333 frames reflects an expanded active working set rather than continuous growth with the complete history.
  }
  \label{fig:runtime_scaling}
\end{figure*}

\textbf{Scaling With Rollout Length.}
The fixed-window design keeps geometry-estimation cost approximately bounded
as the generated trajectory grows.
As shown in Fig.~\ref{fig:runtime_scaling}(a), DA3 inference increases during
the initial warm-up as its input window is filled.
Once the fixed 93-frame window is reached, the inference cost remains
approximately constant as the accumulated rollout grows from 109 to
893 frames.
Thus, the geometry estimator does not repeatedly reprocess the complete
generated history when updating factorized memory.

The memory-side computation shows a similarly bounded trend. As shown in
Fig.~\ref{fig:runtime_scaling}(b), computing the keys and values for projected
references remains nearly constant throughout the steady-state rollout, and
the total history-token preparation cost stabilizes after the native context
is filled. Reference keys and values are shared across the denoising steps of
one generated chunk and recomputed for the next chunk; these K/V tensors are
not retained across chunks.
Pointmap fusion exhibits a discrete workload transition around 333 rollout
frames, where its cost increases as the active fusion working set expands.
Importantly, the cost stabilizes again after this transition rather than
continuing to grow with accumulated rollout length.

\textbf{Runtime Breakdown.}
Table~\ref{tab:runtime_breakdown} reports component-wise latency in the
steady-state regime, measured over chunks 7--56, corresponding to accumulated
rollout lengths from 109 to 893 frames.
Geometry estimation is the dominant cost: the complete DA3 geometry update
requires $54.772\pm1.055$\,s per update.
This quantity includes DA3 inference together with the associated alignment
and geometry-update operations, and is therefore slightly larger than the pure
DA3 inference time reported in Table~\ref{tab:da3_frame_scaling}.

In contrast, preparing the projected reference tokens is substantially cheaper
than geometry estimation. Computing their attention keys and values requires
$1.203\pm0.016$\,s per generation chunk, while preparing the keys and values
for the complete history input, including retained context, takes
$4.985\pm0.057$\,s. These quantities are recomputed for each generated chunk.
Pointmap fusion requires $2.117\pm0.977$\,s on average over the same range.
Its relatively large variance is primarily caused by the workload transition
visible in Fig.~\ref{fig:runtime_scaling}(b): the fusion cost changes from
$0.601\pm0.022$\,s over 109--317 frames to
$2.706\pm0.245$\,s over 333--893 frames, while remaining stable within each
regime.

\begin{table}[t]
    \centering
    \caption{
    Runtime breakdown in the steady-state long-horizon regime.
   Reference K/V computation, highlighted in light green,
    represents the reference-token preparation overhead introduced by
    GROME.
    }
    \label{tab:runtime_breakdown}
    \small
    \setlength{\tabcolsep}{7pt}

    \begin{tabular}{lcc}
        \toprule
        \textbf{Component}
        & \textbf{Mean (s)}
        & \textbf{Std. (s)} \\
        \midrule

        \multicolumn{3}{l}{\textit{Geometry and memory construction}} \\

        DA3 geometry update
        & 54.772
        & 1.055 \\

        Pointmap fusion
        & 2.117
        & 0.977 \\

        \addlinespace[2pt]
        \multicolumn{3}{l}{\textit{History-token preparation}} \\

        Context K/V computation
        & 3.782
        & 0.056 \\

        Reference K/V computation
        & \profilecell{1.203}
        & \profilecell{0.016} \\

        \midrule

        Total history K/V computation
        & 4.985
        & 0.057 \\

        \bottomrule
    \end{tabular}
\end{table}

\textbf{Discussion.}
These results separate two sources of inference overhead in GROME.
The dominant cost arises from external geometric estimation used to construct
factorized memory, whereas preparing projected reference tokens adds a smaller
and stable attention-side overhead. Geometry estimation operates on a fixed
temporal window, while history conditioning consumes references projected into
the target camera views rather than the complete memory bank.
Accordingly, within the evaluated rollout range, neither geometry estimation
nor reference K/V computation exhibits a progressive increase in per-chunk cost with
accumulated history.
Pointmap fusion depends on the active scene geometry and can change when its
working set expands, but it likewise stabilizes within each observed workload
regime and remains substantially cheaper than geometry estimation.
GROME therefore trades additional test-time geometry computation for improved
long-horizon consistency while requiring no training or fine-tuning.

\section{More Results}

\begin{figure*}[t]
  \centering
  \includegraphics[width=\textwidth]
  {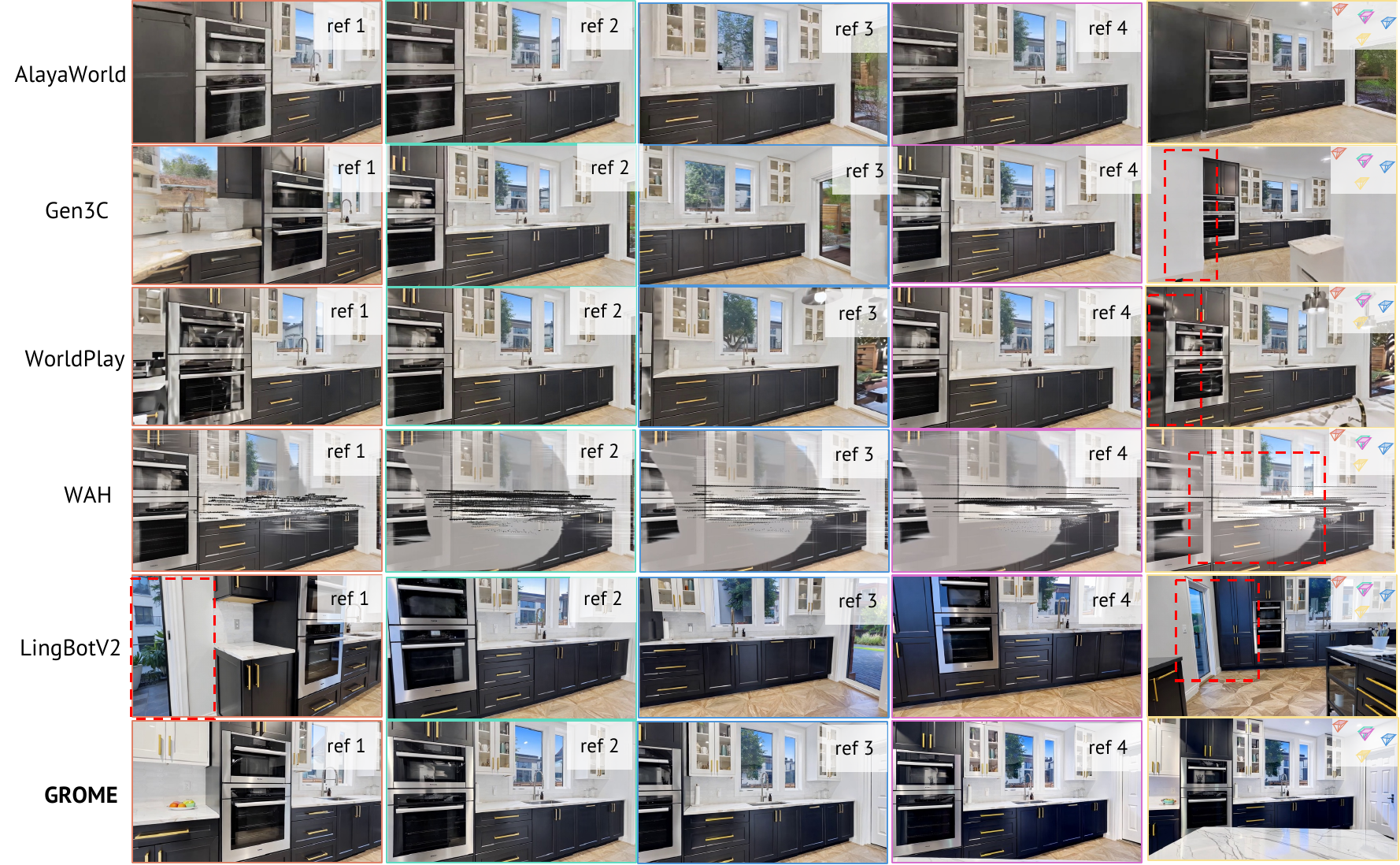}
  \caption{
    \textbf{Static complex-memory revisit-case 1.}
    When exploring both sides of the kitchen, the visual evidence required for the return view is distributed across multiple historical frames, exceeding the model's finite context window. \textbf{GROME} retrieves and consolidates local memories to reconstruct comprehensive, high-fidelity static scene references. Red boxes highlight representative inconsistencies in the baselines.}
  \label{fig:complex_memory_1}
\end{figure*}
\begin{figure*}[t]
  \centering
  \includegraphics[width=\textwidth]
  {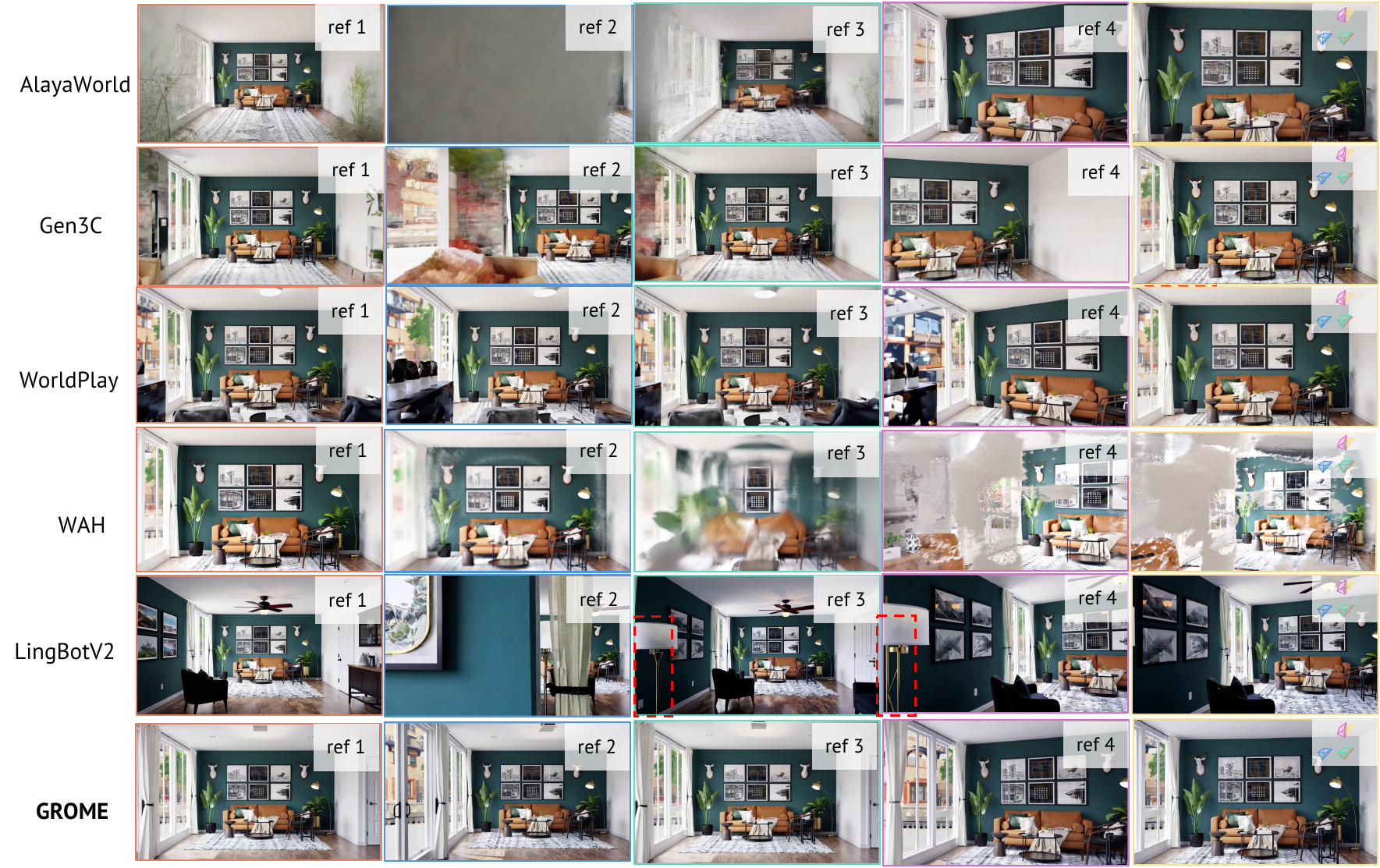}
\caption{
      \textbf{Static complex-memory revisit-case 2.}
      The L-shaped explore--return trajectory distributes pose and scene
      evidence across several views, causing context-based methods to drift or
      omit content. GROME retrieves complementary static scene evidence through
      max-coverage retrieval, preserving the revisit scene structure.
      Red boxes mark representative failures.
      }
  \label{fig:complex_memory_2}
\end{figure*}
\begin{figure*}[t]
  \centering
  \includegraphics[width=\textwidth]
  {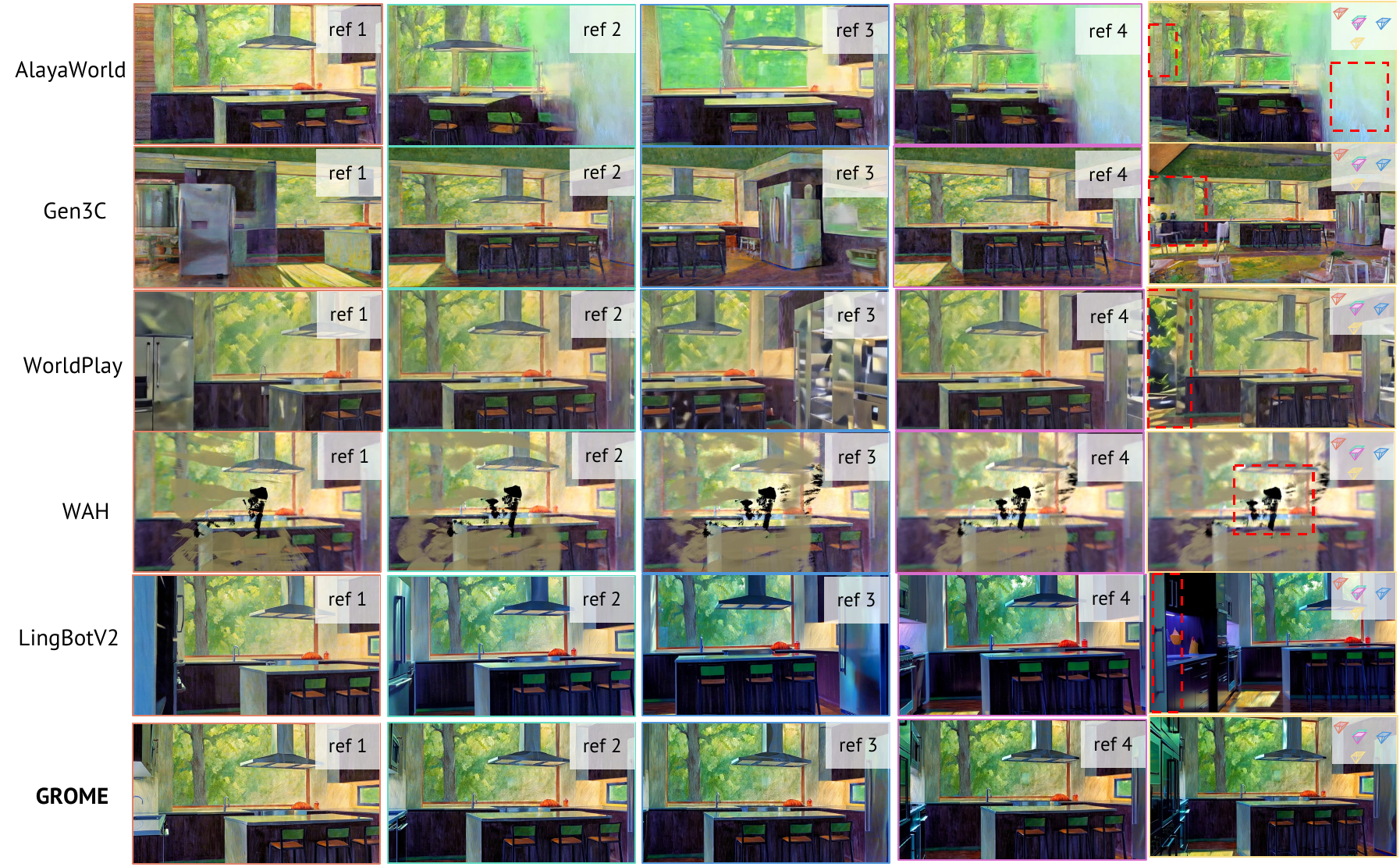}
  \caption{
    \textbf{Static complex-memory revisit-case 3.}
    Complex camera trajectories create multi-view spatial configurations where different parts of the target view are scattered across disparate historical references, exposing the limits of single-frame retrieval. While maintaining a monolithic global cache often suffers from accumulated drift and high overhead, \textbf{GROME} retains historical evidence as local point clouds and selectively retrieves them under target poses. This strategy synthesizes complementary scene evidence to generate the target view with high-fidelity revisit consistency. Red boxes mark representative baseline failures.
    }
  \label{fig:complex_memory_3}
\end{figure*}

\begin{figure*}[t]
  \centering
  \includegraphics[width=\textwidth]{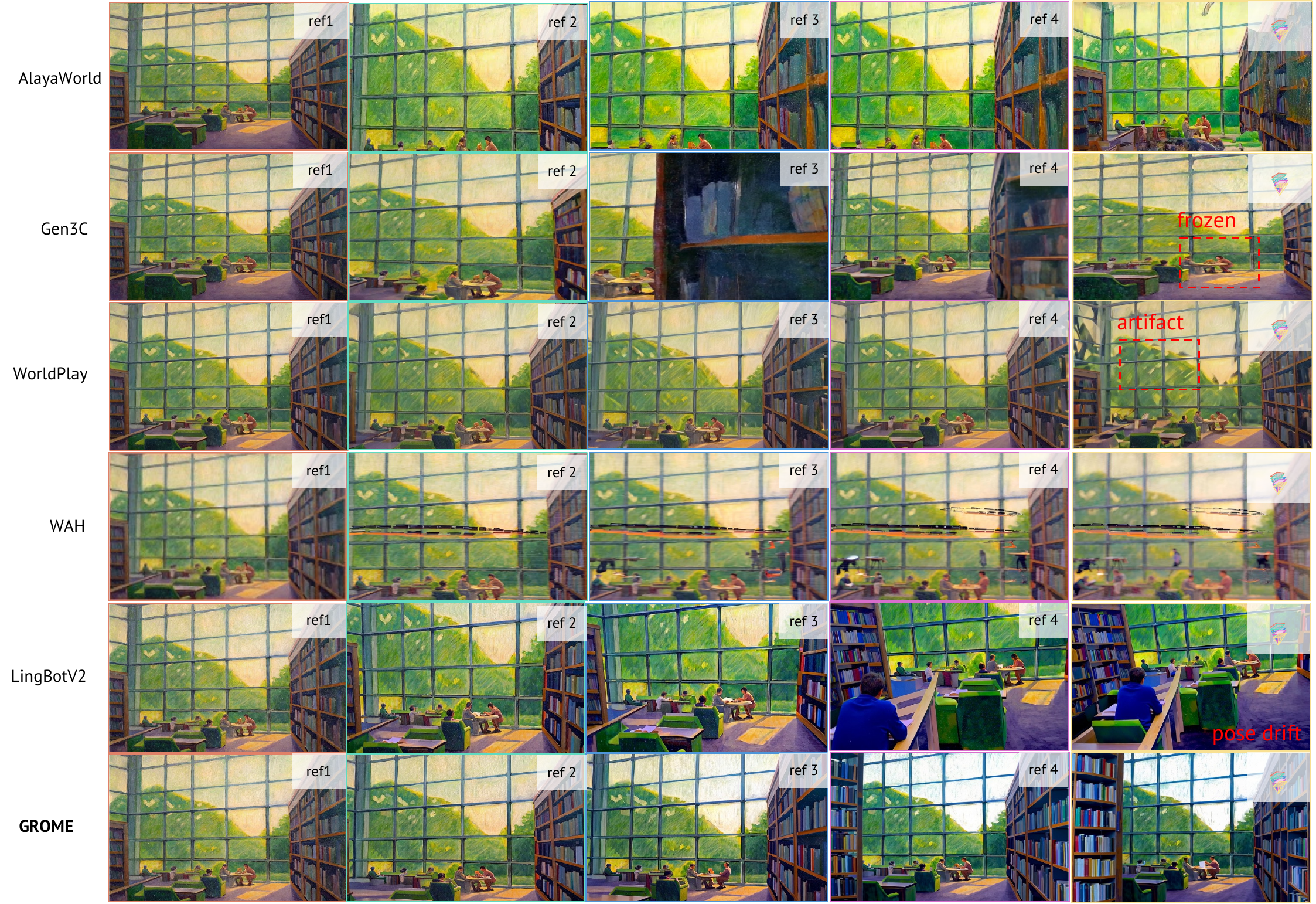}
  \caption{
  \textbf{Dynamic complex-memory revisit-case 1.}
  Explicit-memory baselines overlook dynamic entities during memory construction, freezing them at outdated historical states, whereas context-only generation inevitably loses long-range environmental evidence. By factorizing static structure from dynamic entities, \textbf{GROME} simultaneously preserves background consistency and enables dynamic states to evolve. Red boxes mark representative failures.
  }
  \label{fig:complex_memory_4}
\end{figure*}

\begin{figure*}[t]
  \centering
  \includegraphics[width=\textwidth]{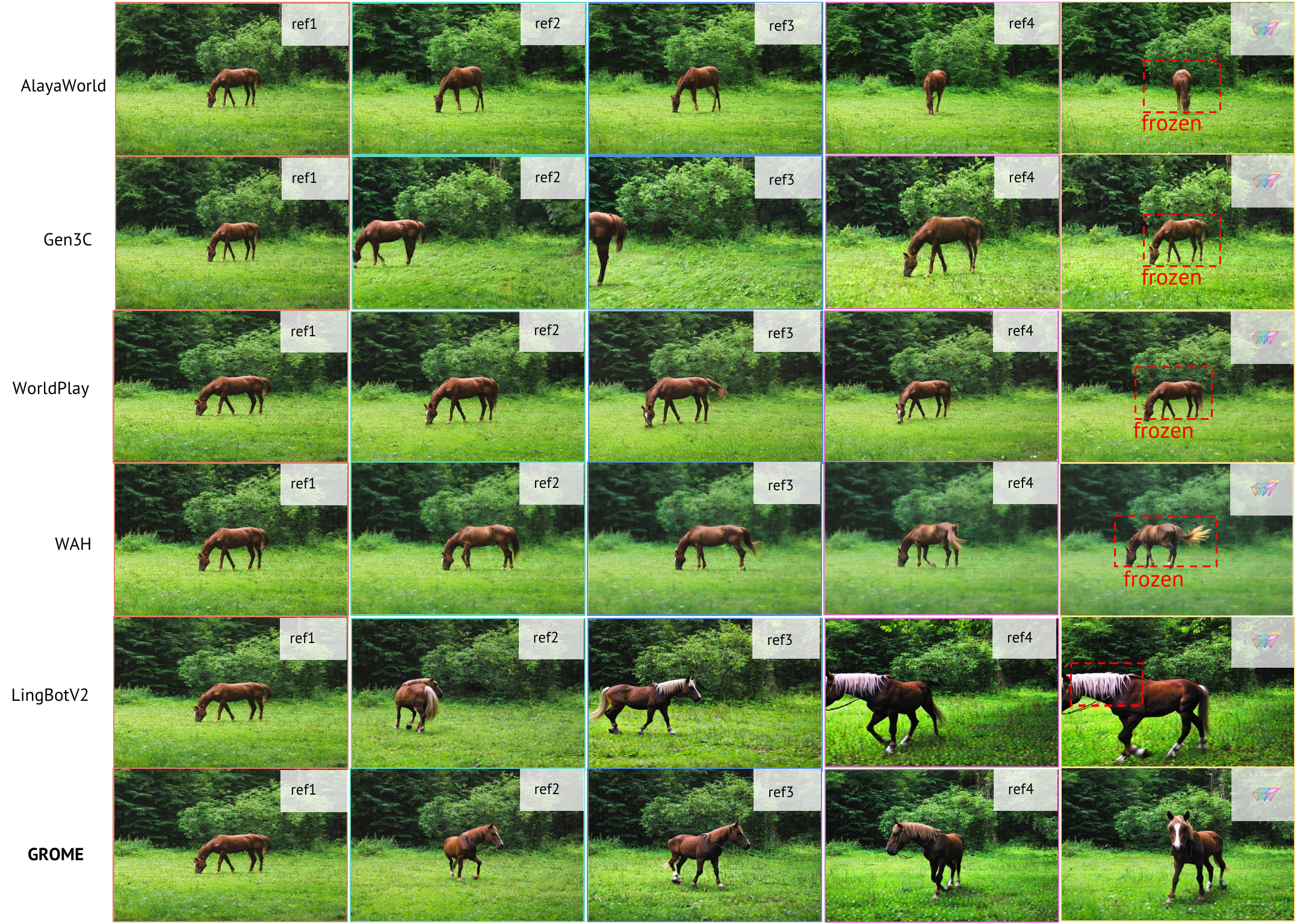}
  \caption{
  \textbf{Dynamic complex-memory revisit-case 2.}
  }
  \label{fig:complex_memory_5}
\end{figure*}

\begin{figure*}[t]
  \centering
  \includegraphics[width=\textwidth]
  {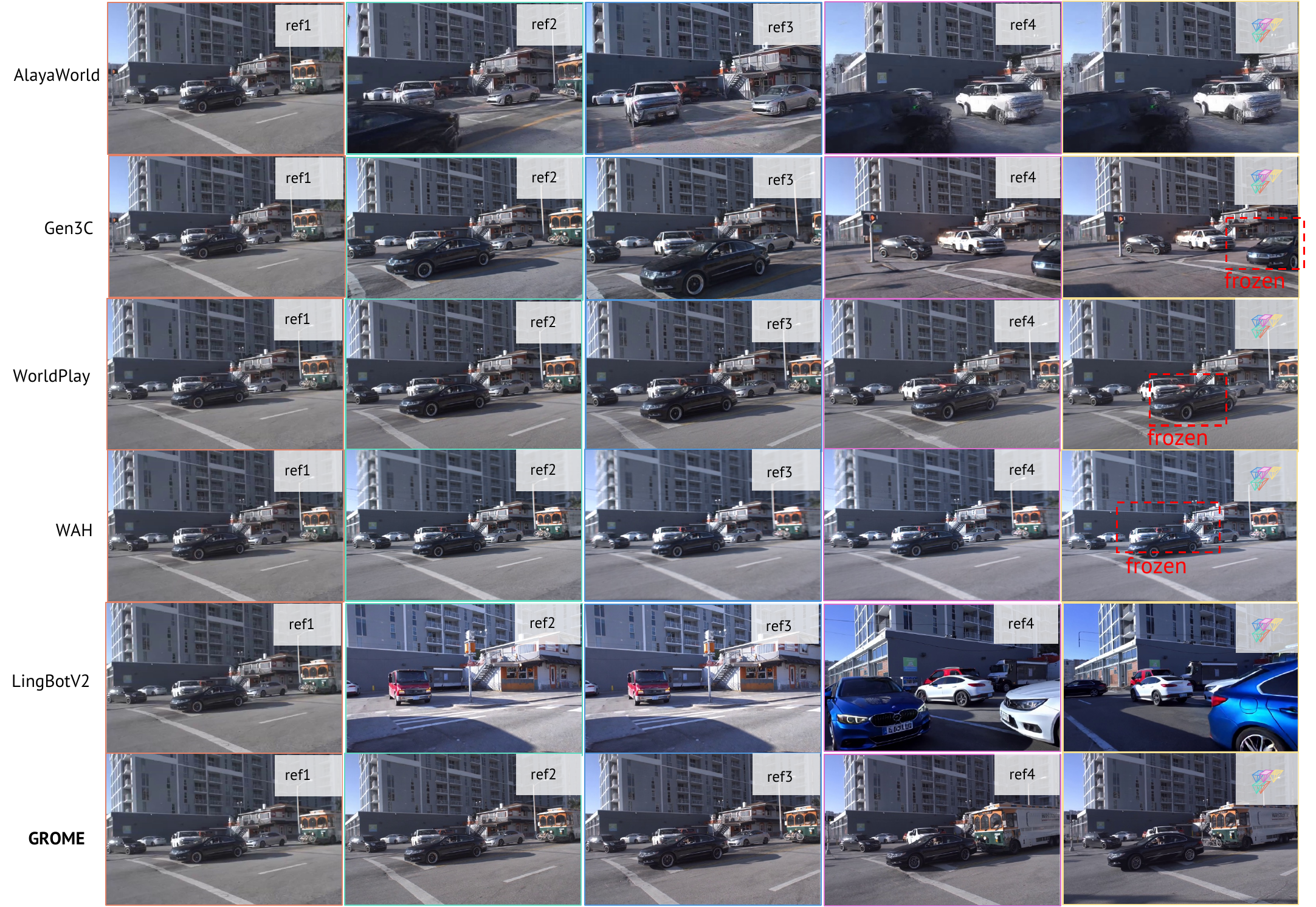}
  \caption{
  \textbf{Dynamic complex-memory revisit-case 3.}
  }
  \label{fig:complex_memory_6}
\end{figure*}
\subsection{Complex-Memory Revisit}
\label{app:complex_memory}

We further evaluate GROME under a more challenging
\emph{complex-memory revisit} setting, where successful long-horizon
generation requires information that cannot be recovered from a single
historical frame.
We consider both \emph{static} and \emph{dynamic} cases, which stress
different aspects of factorized memory.

\textbf{Static Complex-Memory Revisit.}
As shown in Figs.~\ref{fig:complex_memory_1}--\ref{fig:complex_memory_3},
the camera first explores several spatially distinct views
 (\emph{ref 1--4}) before moving to a final target pose.
The content visible from this target view is distributed across multiple earlier
frames, such that no single historical frame provides sufficient
evidence for reconstructing the complete view.
Successful generation therefore requires the model to compose complementary
spatial evidence accumulated throughout the trajectory, rather than merely
retrieve the most similar historical frame.

Across these examples, competing methods frequently preserve only part of
the previously observed environment or introduce local structural
inconsistencies when information from multiple references is required
simultaneously.
Temporal-memory or context-based methods retain direct historical
frames, but their usable spatial coverage remains tied to individual
stored views, making multi-view composition difficult once relevant evidence
is distributed across the trajectory.
Explicit spatial-memory baselines provide stronger geometric grounding, but
can still exhibit incomplete recovery, warping artifacts, or local structural
distortion.
In contrast, GROME stores complementary static content with camera poses in a
shared 3D coordinate system and jointly projects relevant evidence from
multiple local maps.
This allows the final generation to remain compatible with several previously
observed regions simultaneously.

\textbf{Dynamic Complex-Memory Revisit.}
Figs.~\ref{fig:complex_memory_4}--\ref{fig:complex_memory_6} extend the same
setting to scenes containing moving objects.
These examples introduce an additional challenge: historical frames
should not be indiscriminately consolidated, since different references may
contain the same dynamic content at different temporal states.
A successful memory mechanism must therefore combine persistent spatial
evidence from the scene while resolving dynamic evidence according to its
recency within the current target view.

This distinction exposes characteristic failure modes of existing memory
strategies.
Methods that spatially preserve historical frames without explicitly
separating static and dynamic content can lock moving objects to stale
states, resulting in the frozen motion visible in
Figs.~\ref{fig:complex_memory_4} and~\ref{fig:complex_memory_6}.
Other methods exhibit accumulated rendering artifacts, pose drift, or
long-horizon changes in scene content.
GROME instead applies different reuse rules to the two components of one
factorized memory: the static component combines complementary spatial content
across frames, whereas the dynamic component retains source times and resolves
overlapping projected content by recency.
Consequently, GROME can preserve the persistent world structure without
forcing moving objects to match an outdated historical frame.

Because different world models synthesize different intermediate content,
their final views are not pixel-wise aligned to a common generated target.
We therefore use these examples as qualitative comparisons rather than
image-level quantitative evaluations.
Together, the six cases demonstrate that complex long-horizon memory requires
not only broader spatial coverage, but also different reuse strategies for
persistent scene structure and time-varying visible content.

\clearpage
\begin{figure*}[p]
  \centering

  \includegraphics[
      width=\textwidth,
      height=0.36\textheight,
      keepaspectratio
  ]{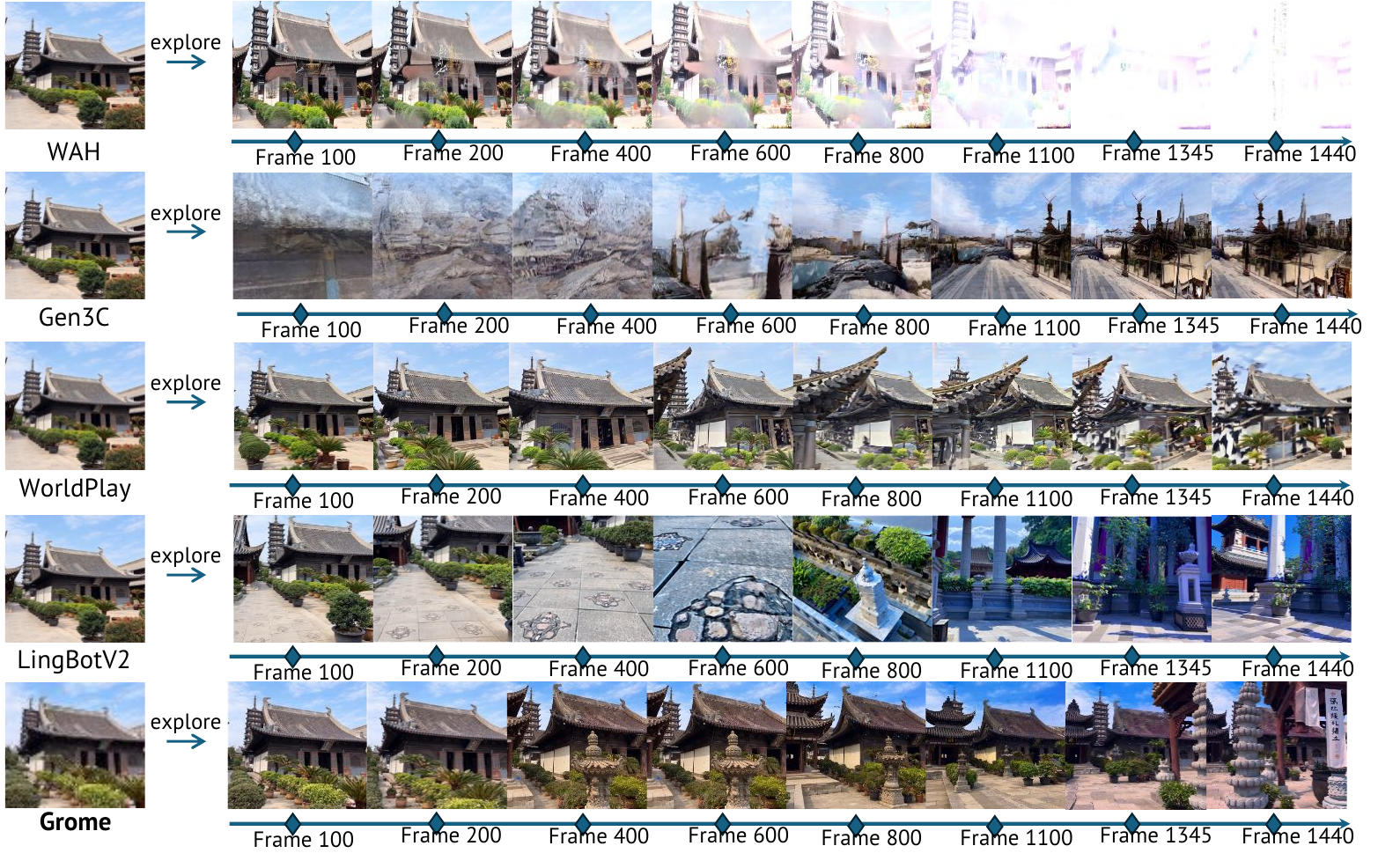}

  \vspace{-0.4em}
  \centerline{\small (a) Static scene}

  \vspace{0.6em}

  \includegraphics[
      width=\textwidth,
      height=0.36\textheight,
      keepaspectratio
  ]{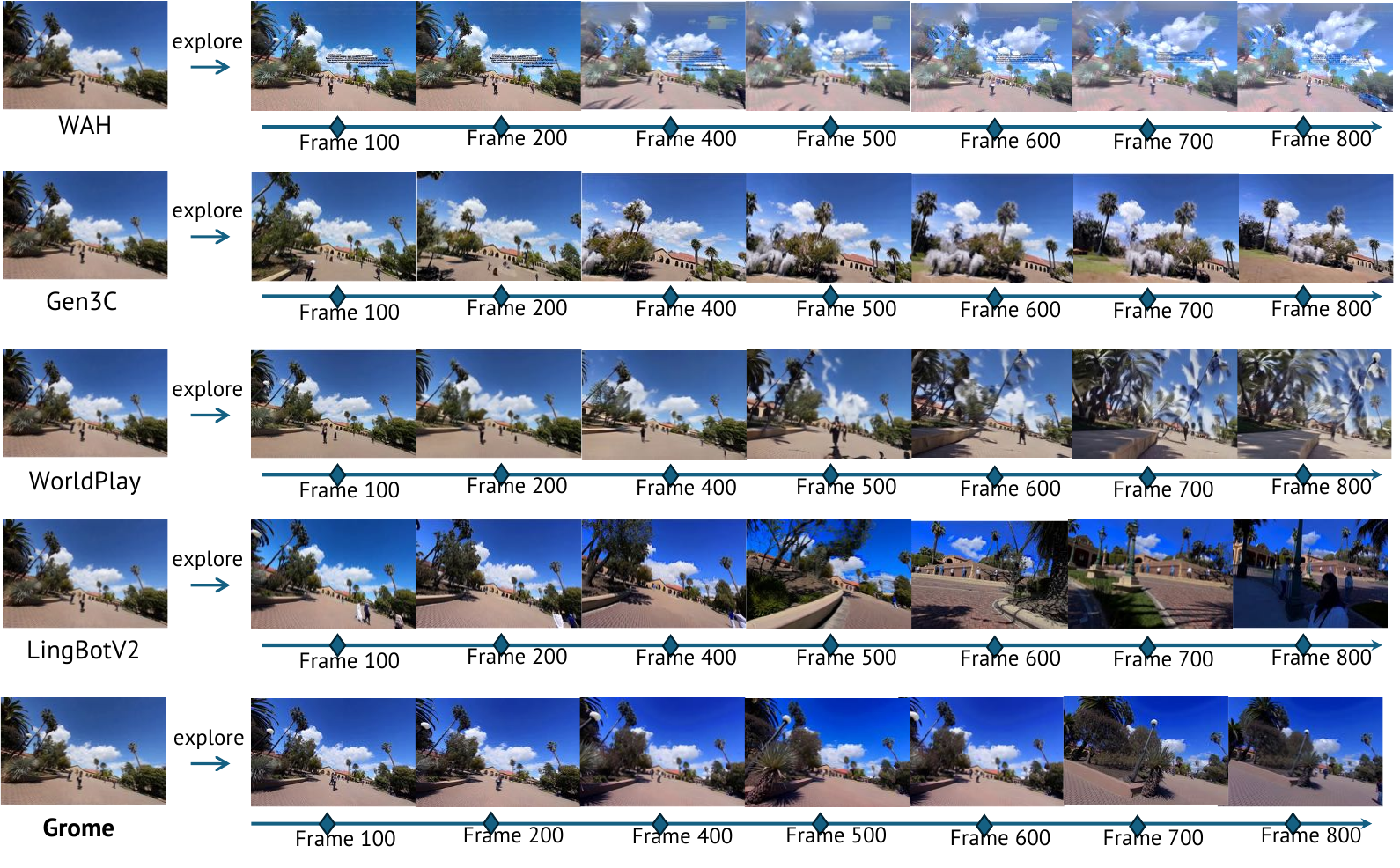}

  \vspace{-0.4em}
  \centerline{\small (b) Dynamic scene}
  \caption{
  \textbf{Long-horizon qualitative comparison.}
  When historical views leave the finite context, implicit memory and explicit memory-based
  baselines drift from the scene, while GROME preserves
  static structure beyond 1,400 frames. On dynamic scenes, factorized memory
  retains evolving dynamic entities without forcing them to remain in a stale
  state.
  }
  \label{fig:longhorizon}
\end{figure*}

\subsection{Long-Horizon Qualitative Comparison}
\label{app:long_horizon_comparison}

Our main quantitative evaluation uses trajectories of 15--18\,s
(360--432 frames at 24 FPS). Here, we additionally evaluate substantially
longer autoregressive rollouts exceeding 800 frames to examine
failure modes that become apparent only over extended horizons. Since such
long rollouts involve progressively accumulated generation errors and do not
admit a stable frame-wise quantitative comparison throughout the entire
sequence, we report representative qualitative results instead.
These examples examine whether different methods can preserve previously generated spatial structure and appearance after the corresponding frames
have long left the model's effective context window.

\textbf{Static Scenes.}
Fig.~\ref{fig:longhorizon}(a) compares representative rollouts over more than
1,400 frames. Vanilla LingBotV2 gradually loses the geometry and appearance of
the initial environment, eventually generating views that are only weakly
related to the observed scene. Temporal or warp-based baselines alleviate
short-term drift, but their errors continue to accumulate over longer
horizons. WAH exhibits severe appearance degradation in the later stages,
while WorldPlay progressively distorts scene structure under continued camera
motion. Gen3C preserves stronger local geometry, but its generated views still
show noticeable long-horizon appearance and structural changes. In contrast,
GROME produces views that remain closer to the characteristic architecture,
layout, and appearance of the earlier generated environment, including at very
late frames. These qualitative examples suggest that pose-indexed historical
content provides a useful anchor beyond recent or retrieved frames.

\textbf{Dynamic Scenes.}
Long-horizon consistency becomes more challenging when the environment also
contains dynamic content. As shown in Fig.~\ref{fig:longhorizon}(b), simply
preserving an early frame is insufficient: the model must retain the
underlying scene while allowing visible moving content to change.
LingBotV2 exhibits substantial scene and appearance drift as the rollout
progresses, while temporal-memory baselines gradually deviate from the initial
environment despite remaining locally plausible. WAH further accumulates
noticeable appearance degradation at later frames. Gen3C preserves the static
scene structure relatively well, but moving objects progressively disappear
from the generated world, indicating that strong scene consistency can come at
the expense of dynamic preservation. In contrast, GROME maintains the
recognizable scene layout and moving content throughout the rollout. The
backbone, rather than factorized memory, generates future motion. This behavior
is consistent with the quantitative results, where GROME improves revisit
similarity while maintaining a Motion Accuracy score comparable to the frozen
backbone.

Overall, these examples complement the quantitative revisit evaluation by
showing how errors accumulate over time. GROME conditions each generated chunk
on projected content from factorized memory, delaying the
scene and appearance drift observed in competing methods.

\subsection{Additional Application Cases}
\label{app:applications}

This section expands the two applications summarized in
Sec.~\ref{sec:applications}. Both use the inference pipeline described in the
main paper; they differ in how factorized memory is initialized and in the camera
trajectory used for the subsequent rollout.

\textbf{Real-World Video Continuation.}
\label{app:real_world_continuation}
\begin{figure}[t]
  \centering
  \includegraphics[width=0.9\linewidth]{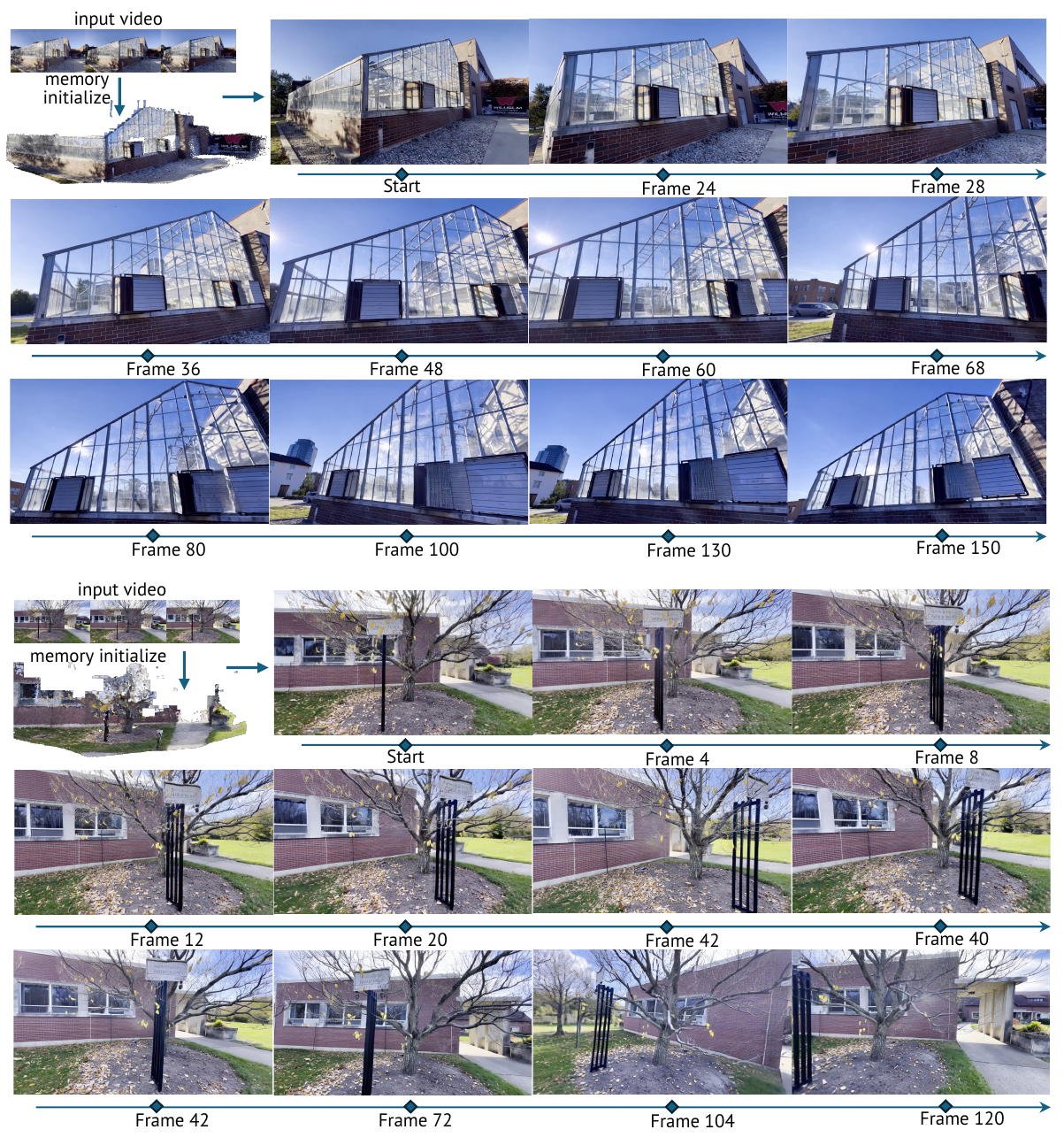}
  
  \caption{\textbf{Video-to-Memory Continuation.}
  Additional real-world video-continuation cases. Each row pairs an
  observed video with a GROME rollout generated along a new camera
  trajectory. Revisit frames assess retention of previously observed scene
  content, while novel-view frames show the continuation beyond the input.}
  \label{fig:appendix_real_world_continuation}
\end{figure}
The cases in Fig.~\ref{fig:appendix_real_world_continuation} are organized to
separate two aspects of continuation: recovering scene evidence when the
camera revisits an observed region and generating views outside the recorded
trajectory. Each case places the input frames before the rollout, enabling a
direct comparison of retained appearance and layout with the source video.

\label{app:spatial_exploration}
\begin{figure}[t]
  \centering
  \includegraphics[width=0.9\linewidth]{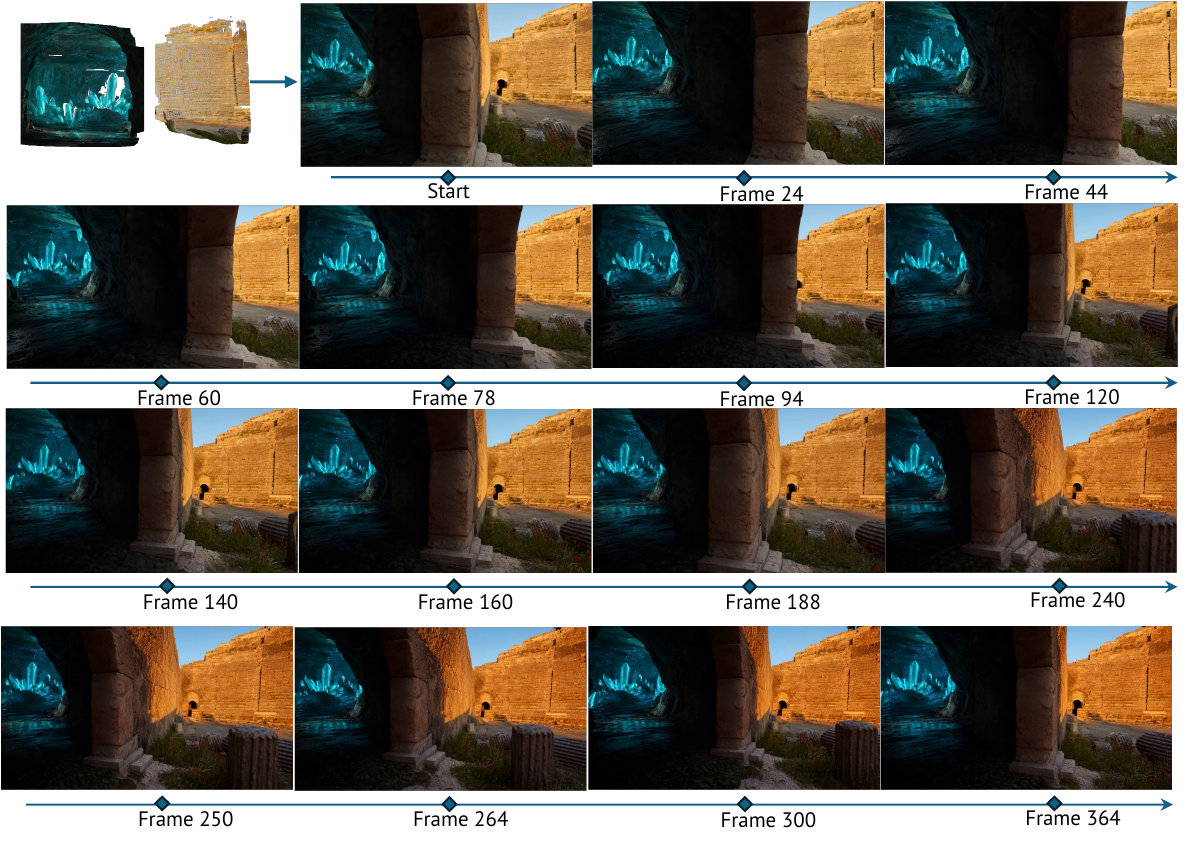}
  \caption{\textbf{Spatially Grounded Exploration.}
  Additional spatially grounded exploration cases. Each row shows
  the source evidence, the memory transformation or composition, and the
  resulting rollout under a novel camera trajectory. This arrangement exposes
  which visual content originates from each memory region.}
  \label{fig:appendix_spatial_exploration}
\end{figure}
\textbf{Spatially Guided Exploration.}
Figure~\ref{fig:appendix_spatial_exploration} uses the memory operation as the
reference point for each case. Showing the source frames beside the
edited memory and rollout makes it possible to assess whether the generated
views follow the requested spatial configuration while retaining the
appearance of the contributing regions.
\clearpage
\subsection{Robustness to Imperfect Geometric Projection}
\begin{figure*}[t]
  \centering
  \includegraphics[width=0.95\textwidth]{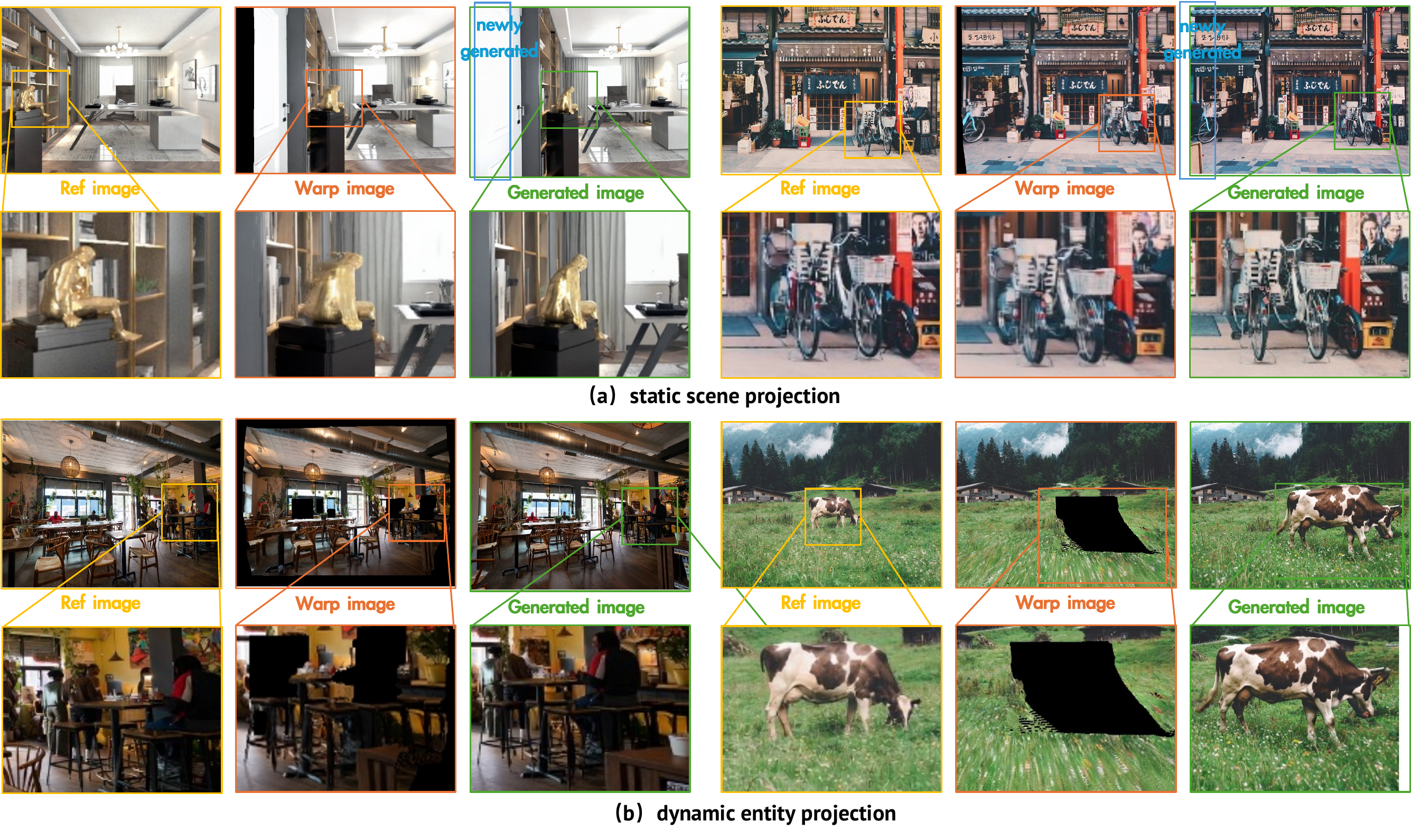}
  \caption{
  \textbf{Robustness to imperfect geometric projection.}
  Although rendered historical memories often exhibit geometric artifacts such as disocclusion holes and blur, \textbf{GROME} leverages them as soft spatial guidance rather than rigid pixel-level targets via timestep-aware memory routing. Guided by pretrained world priors, the frozen model corrects moderate rendered distortions and plausibly synthesizes unobserved regions during finer denoising steps while maintaining scene consistency.
  }
  \label{fig:depth_robustness}
\end{figure*}
GROME relies on explicit geometric projection to bring historical content into the target camera view and therefore, similar to other 3D-conditioned world models~\citep{ren2025gen3c}, is affected by the quality of depth and
camera estimation. Importantly, however, the projected memory in GROME is used as a reference rather than a pixel-wise reconstruction target. As shown in Fig.~\ref{fig:depth_robustness}, the warped references may contain misalignment, blur, missing regions, or occlusion-induced holes, while the generated results recover coherent object appearance and local scene structure instead of directly reproducing these artifacts.

This robustness comes from the interaction between geometric memory and the pretrained world prior. Projection provides spatially aligned evidence about previously observed content, while the frozen video world model remains free to reconcile imperfect references with the current context and synthesize content where the projected evidence is unreliable or unavailable.
Validity-aware masking further prevents invalid
memory regions from entering attention. Consequently,
moderate geometric errors typically degrade the quality of the guidance
rather than being copied directly into the generated frame.

\section{Limitations and Failure Cases}
\label{app:limitations}
\begin{figure*}[t]
  \centering

  \begin{minipage}[t]{0.49\textwidth}
      \centering
      \includegraphics[width=\linewidth]
      {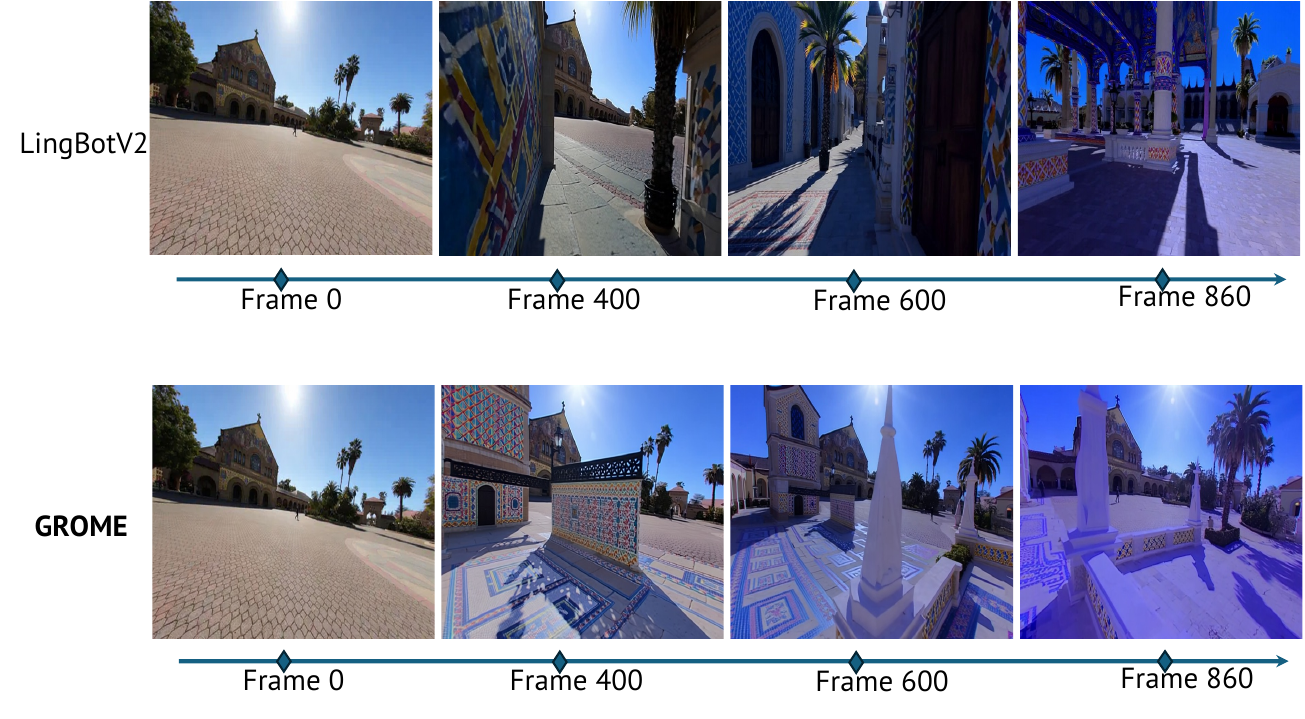}
      \vspace{-0.4em}
      \centerline{\small (a) Backbone-dependent appearance drift}
  \end{minipage}
  \hfill
  \begin{minipage}[t]{0.49\textwidth}
      \centering
      \includegraphics[width=\linewidth]
      {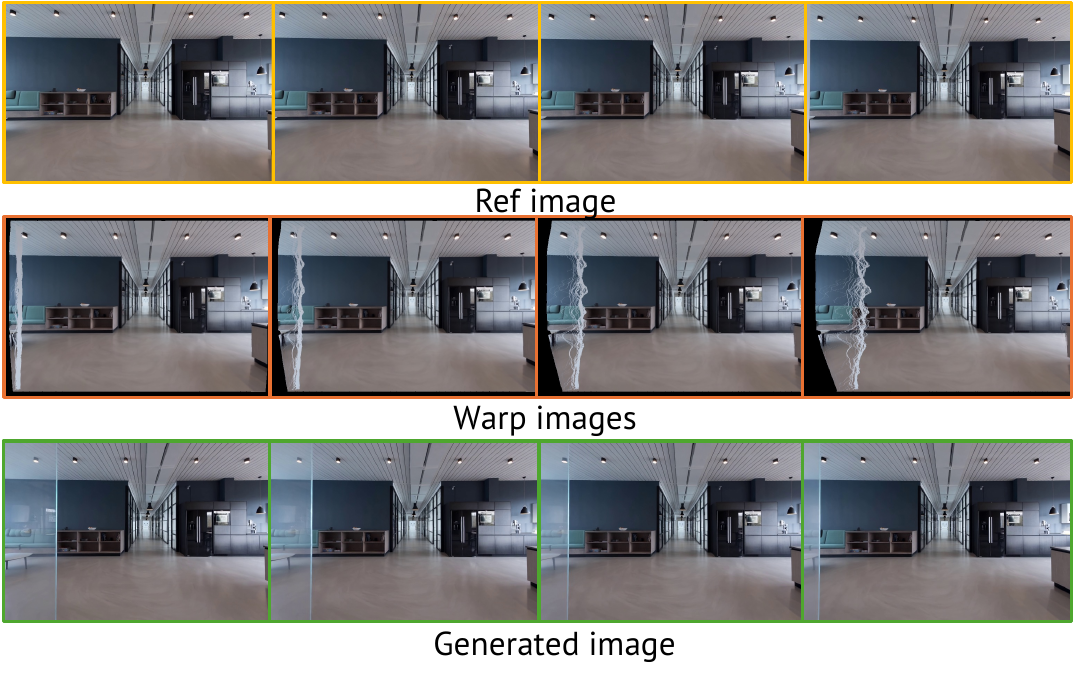}
      \vspace{-0.4em}
      \centerline{\small (b) Depth-induced projection artifacts}
  \end{minipage}

  \caption{
  \textbf{Representative failure cases of GROME.}
  (a) At long rollout horizons, appearance and color drift inherited
  from the frozen backbone can still accumulate despite Reference Value Adaptation.
  (b) Severe local geometry errors can produce sparse or corrupted warped
  evidence. Although the pretrained VWM usually tolerates moderate projection
  imperfections, extreme artifacts may be interpreted as plausible scene
  content and propagate into the generated frame.
  }
  \label{fig:failure_cases}
\end{figure*}

As a training-free framework, GROME remains bounded by both the generative
capability of the frozen backbone and the quality of the external geometric
estimation. As shown in Fig.~\ref{fig:failure_cases}(a), LingBotV2 gradually
accumulates appearance and color drift over very long autoregressive rollouts.
Although Reference Value Adaptation alleviates this mismatch by aligning
historical reference values with the current rollout, GROME cannot eliminate such
backbone-intrinsic degradation, and residual drift can still become noticeable
at long horizons. GROME is also tolerant to moderate depth and
projection errors, as the pretrained VWM can often correct local artifacts
using its learned world prior. However, under extreme geometric failures,
this self-correction can break down. In Fig.~\ref{fig:failure_cases}(b), a
small foreground region becomes poorly reconstructed during the rollout,
producing sparse and erroneous points in the warped reference. The frozen
video model subsequently interprets these artifacts as plausible translucent
or glass-like structures, leading to incorrect generation in the affected
region. These cases highlight that GROME improves the reuse of historical
evidence without retraining the generator, but does not remove the fundamental
uncertainties inherited from either the backbone or geometric estimation.
 Future work could move beyond the current training-free setting by jointly learning how to condition
the generative backbone on projected references, allowing the model
to better recognize and correct imperfect geometric guidance rather than
relying solely on the pretrained world prior.

\end{document}